\documentclass[10pt,twocolumn,letterpaper]{article}

\usepackage{iccv}
\usepackage{times}
\usepackage{epsfig}
\usepackage{graphicx}
\usepackage{amsmath}
\usepackage{amssymb}

\usepackage{microtype}
\usepackage{subfigure}
\usepackage{booktabs} % for professional tables
\usepackage{multirow}
\usepackage{stackengine}
\usepackage[table]{xcolor}

\usepackage[pagebackref=true,breaklinks=true,letterpaper=true,colorlinks,bookmarks=false]{hyperref}

\newcommand{\minisection}[1]{\vspace{0mm}\noindent{\textbf{#1.}}}
\newcommand{\ignore}[1]{}

\iccvfinalcopy % *** Uncomment this line for the final submission

\def\iccvPaperID{5138} % *** Enter the ICCV Paper ID here

\begin{document}

%%%%%%%%% TITLE
\title{Toward Understanding Catastrophic Forgetting in Continual Learning}

\author{Cuong V.~Nguyen$^\dagger$, Alessandro Achille$^\dagger$, Michael Lam$^\dagger$, Tal Hassner$^\ddagger$\thanks{Work done at Amazon.}, \\ Vijay Mahadevan$^\dagger$, Stefano Soatto$^\dagger$\\[10pt]
$^\dagger$Amazon Web Services \\
{\tt \small \{nguycuo,aachille,michlam,vmahad,soattos\}@amazon.com} \\[3pt]
$^\ddagger$Facebook Inc.\\
{\tt \small \{talhassner\}@gmail.com}
}

\maketitle

\begin{abstract}
We study the relationship between catastrophic forgetting and properties of task sequences. In particular, given a sequence of tasks, we would like to understand which properties of this sequence influence the error rates of continual learning algorithms trained on the sequence. To this end, we propose a new procedure that makes use of recent developments in task space modeling as well as correlation analysis to specify and analyze the properties we are interested in. As an application, we apply our procedure to study two properties of a task sequence: (1) total complexity and (2) sequential heterogeneity. We show that error rates are strongly and positively correlated to a task sequence's total complexity for some state-of-the-art algorithms. We also show that, surprisingly, the error rates have no or even negative correlations in some cases to sequential heterogeneity. Our findings suggest directions for improving continual learning benchmarks and methods.
\end{abstract}

\section{Introduction}
\label{sec:intro}

Continual learning (or life-long learning)~\cite{ring1997child,schlimmer1986case,sutton1993online} is the ability of a machine learning model to continuously learn from a stream of data, which could possibly be non-iid or come from different but related tasks. A continual learning system is required to adapt its current model to the new tasks or datasets without revisiting the previous data. Such a system should be able to positively transfer its current knowledge (summarized in its model) to the new tasks using as few data as possible, to avoid catastrophically forgetting the old tasks, and to transfer back its knowledge from new tasks to old tasks in order to improve overall performance.

In recent years, interest in continual learning has risen~\cite{achille2018life, kirkpatrick2017overcoming, li2018learning, lopez2017gradient, nguyen2018variational, schwarz2018progress, serra2018overcoming, shin2017continual, zenke2017continual}, especially from the deep learning research community, due to its potential to reduce training time and training set sizes (e.g., by continuously adapting from previous models), both of which are critical to the training of modern deep neural networks. Solving continual learning is also an essential step toward artificial general intelligence as it allows machines to continuously adapt to changes in the environment with minimal human intervention, a process analogous to human learning.

However, continual learning by deep models has proven to be very challenging due to {\em catastrophic forgetting}, a long known problem of training deep neural networks~\cite{ans1997avoiding, ans2000neural, french1999catastrophic,  goodfellow2013empirical, mccloskey1989catastrophic, ratcliff1990connectionist, robins1995catastrophic}. Catastrophic forgetting refers to the tendency of a model to forget all its previously learned tasks if not trained properly on a new task, e.g., when fine-tuning on the new task for a long time without proper regularization to the previous model parameters.

Recent work attempted to tackle this problem either by better training algorithms~\cite{kirkpatrick2017overcoming, lee2017overcoming, ritter2018online, zenke2017continual}, structure sharing~\cite{rusu2016progressive, sharif2014cnn, yoon2018lifelong}, episodic memory~\cite{chaudhry2019efficient, lopez2017gradient, nguyen2018variational}, machine-generated pseudo-data~\cite{isele2018selective, li2018learning, shin2017continual}, or a combination of these approaches~\cite{nguyen2018variational, schwarz2018progress}. Benchmarks to compare these methods typically constructed a sequence of tasks and then measured the algorithms' performance when transferring from one task to another. Two popular examples of these benchmarks are the {\em permuted MNIST}~\cite{goodfellow2013empirical} and {\em split MNIST}~\cite{zenke2017continual}.

In this paper, we seek to understand catastrophic forgetting at a more fundamental level. Specifically, we investigate the following question:

\vspace{1mm}
\emph{Given a sequence of tasks, which properties of the tasks influence the hardness of the entire sequence?}
\vspace{1mm}

We measure {\em task sequence hardness} by the final error rate of a model trained sequentially on the tasks in the sequence.

An answer to this question is useful for continual learning research in several ways. First, it helps us estimate the hardness of a benchmark based on its individual tasks, thereby potentially assisting the development of new and better benchmarks for continual learning. Additionally, knowing the hardness of a task sequence allows us to estimate a priori the cost and limits of running continual learning algorithms on it. Crucially, by gaining a better understanding of catastrophic forgetting at a more fundamental level, we gain more insights to develop better methods to mitigate it.

This work is the first attempt to answer the above question. We propose a new and general procedure that can be applied to study the relationship between catastrophic forgetting and properties of task sequences. Our procedure makes use of recent developments in task space modeling methods, such as the \emph{Task2Vec} framework~\cite{achille2019task2vec}, to specify the interested properties. Then, we apply correlation analysis to study the relationship between the specified properties and the actual measures of catastrophic forgetting.

As an application, we use our procedure to analyze two properties of a task sequence---\emph{total complexity} and \emph{sequential heterogeneity}---and design experiments to study their correlations with the sequence's actual hardness. We refer to total complexity as the total hardness of individual tasks in the sequence, while sequential heterogeneity measures the total dissimilarity between pairs of consecutive tasks.

We show how these two properties are estimated using the Task2Vec framework~\cite{achille2019task2vec}, which maps datasets (or equivalently, tasks) to vectors on a vector space. We choose these two properties for our analysis because of their intuitive relationships to the hardness of task sequences: since continual learning algorithms attempt to transfer knowledge from one task to another, both the hardness of each individual task and the dissimilarity between them should play a role in determining the effectiveness of the transfer.

The findings from our analysis are summarized below. 

\begin{itemize}
\item Total complexity has a \emph{strong correlation} with the task sequence hardness measured by the actual error rate.
\item Sequential heterogeneity has \emph{little or no correlation} with the task sequence hardness. When factoring out the task complexity, we even find negative correlations in some cases.
\end{itemize}

The first finding, although expected, emphasizes that we should take into account the complexity of each task when designing new algorithms or benchmarks, which is currently lacking in continual learning research. Besides, the research community is currently somewhat divided on the issue whether task similarity helps or hurts continual learning performance. Some authors showed that task similarity helps improve performance in the context of transfer learning~\cite{achille2019task2vec, ammar2014automated, ruder2017learning}, while some others conjectured that task dissimilarity could help improve continual learning performance~\cite{farquhar2018towards}. Our second finding gives evidence that supports the latter view.

Deeper analysis into these phenomena suggests that (a) the task sequence hardness also depends on the ability to backward transfer (i.e., learning a new task helps a previous task) and (b) continual learning algorithms should be customized for specific task pairs to improve their effectiveness. We give detailed analysis and discussions in Sec.~\ref{sec:discuss}.

\section{Continual learning algorithms and existing benchmarks}
\label{sec:cl}

We overview modern continual learning algorithms and existing benchmarks used to evaluate them. For more comprehensive reviews of continual learning, we refer to Chen et al.~\cite{chen2016lifelong} and Parisi et al.~\cite{parisi2018continual}

\subsection{Continual learning algorithms}

The simplest and most common approaches to continual learning use weight regularization to prevent catastrophic forgetting. Weight regularization adds a regularizer to the likelihood during training to pull the new weights toward the previous weights. It has been improved and applied to continual learning of deep networks in the elastic weight consolidation (EWC) algorithm~\cite{kirkpatrick2017overcoming}, where the regularizer is scaled by the diagonal of the Fisher information matrix computed from the previous task. Since the diagonal Fisher information approximates the average Hessian of the likelihoods, EWC is closely related to Laplace propagation~\cite{huszar2018note, eskin2004laplace}, where Laplace's approximation is applied after each task to compute the regularizers. Besides Fisher information, the path integral of the gradient vector field along the parameter optimization trajectory can also be used for the regularizer, as in the synaptic intelligence (SI) approach~\cite{zenke2017continual}.

Another form of regularization naturally arises by using Bayesian methods. For instance, variational continual learning (VCL)~\cite{nguyen2018variational, swaroop2018improving} applied a sequence of variational approximations to the true posterior and used the current approximate posterior as prior for the new task. The Kullback-Leibler term in the variational lower bound of VCL naturally regularizes the approximate posterior toward the prior. Improved training procedures have also been developed for this type of approximate Bayesian continual learning through the use of natural gradients~\cite{chen2018facilitating, tseran2018natural}, fixed-point updates~\cite{zeno2018task}, and local approximation~\cite{bui2018partitioned}. More expressive classes of variational distributions were also considered, including channel factorized Gaussian~\cite{kochurov2018bayesian}, multiplicative normalizing flow~\cite{kochurov2018bayesian}, or structured Laplace approximations~\cite{ritter2018online}.

The above methods can be complemented by an episodic memory, sometimes called a {\em coreset}, which stores a subset of previous data. Several algorithms have been developed for utilizing coresets, including gradient episodic memory~(GEM) \cite{lopez2017gradient}, averaged GEM~\cite{chaudhry2019efficient}, coreset VCL~\cite{nguyen2018variational}, and Stein gradients coreset~\cite{chen2018facilitating}.

Other algorithmic ideas to prevent catastrophic forgetting include moment matching~\cite{lee2017overcoming}, learning without forgetting~\cite{li2018learning}, and deep generative replay~\cite{farquhar2018towards, kamra2017deep, shin2017continual}. Structure sharing~\cite{rusu2016progressive, schwarz2018progress} is also another promising direction that can be combined with the above algorithmic solutions to improve continual learning.

\subsection{Existing benchmarks}\label{sec:existingdb}

The most common benchmarks for continual learning use MNIST~\cite{lecun2010mnist} as the base dataset and construct various task sequences for continual learning. For example, permuted MNIST~\cite{goodfellow2013empirical} applies a fixed random permutation on the pixels of MNIST input images for each task, creating a sequence of tasks that keep the original labels but have different input structures. Split MNIST~\cite{zenke2017continual}, on the other hand, considers five consecutive binary classification tasks based on MNIST: 0/1, 2/3, \ldots, 8/9. Another variant is rotated MNIST~\cite{lopez2017gradient}, where the digits are rotated by a fixed angle between 0 and 180 degrees in each task. Similar constructions can also be applied to the not-MNIST set~\cite{notMNIST}, the fashion MNIST set~\cite{xiao2017fashion}, or the CIFAR set~\cite{krizhevsky2009learning} such as in the split not-MNIST~\cite{nguyen2018variational} and split CIFAR benchmarks \cite{zenke2017continual}.

Other continual learning benchmarks include ones typically used for reinforcement learning. For instance, Kirkpatrick et al. \cite{kirkpatrick2017overcoming} tested the performance of EWC when learning to play Atari games. Schwarz et al. \cite{schwarz2018towards} proposed a new benchmark for continual learning based on the StarCraft II video game, where an agent must master a sequence of skills without forgetting the previously acquired skills.

\section{Analysis of catastrophic forgetting}
\label{sec:analysis}

Recent developments in task space modeling, such as Task2Vec~\cite{achille2019task2vec} and Taskonomy~\cite{zamir2018taskonomy}, provide excellent tools to specify and analyze relationships between different tasks from data. In this paper, we propose a novel and general procedure that utilizes these tools to study catastrophic forgetting. Our procedure is conceptually simple and can be summarized in the following steps:
\begin{enumerate}
\item \emph{Specify the properties} of a task sequence that we are interested in and \emph{estimate these properties} using a suitable task space modeling methods.
\item \emph{Estimate actual measures of catastrophic forgetting} from real experiments. In our case, we measure catastrophic forgetting by the task sequence hardness, defined as the final error rate of a model trained sequentially on the sequence.
\item Use \emph{correlation analysis} to study the correlations between the estimated properties in Step 1 and the actual measures in Step 2.
\end{enumerate}

This procedure can be used even in other cases, such as transfer or multi-task learning, to study properties of new algorithms. For the rest of this paper, we demonstrate its use for analyzing two properties of task sequences and their effects on continual learning algorithms.

\section{Total complexity and sequential heterogeneity of task sequences}
\label{sec:task_eval}

We define two properties that we would like to investigate: the \emph{total complexity} and \emph{sequential heterogeneity} of a task sequence, and detail the methodology used to estimate these quantities from data. We start by first introducing the Task2Vec framework~\cite{achille2019task2vec}, the main tool that we employ to quantify the above properties.

\subsection{Preliminaries: Task2Vec}
\label{sec:task2vec}

Task2Vec~\cite{achille2019task2vec} is a recently developed framework for embedding visual classification tasks as vectors in a real vector space. The embeddings have many desirable properties that allow reasoning about the semantic and taxonomic relations between different visual tasks. This is one of several recent attempts to provide tools for understanding the structure of task space. Other related efforts that can be used as alternatives to Task2Vec include, e.g., \cite{edwards2016towards,tran2019transferability,zamir2018taskonomy}.

Given a labeled classification dataset, $\mathcal{D} = \{ (x_i, y_i) \}_{i=1}^N$, Task2Vec works as follows. First, a network pre-trained on a large dataset (e.g., ImageNet), called the probe network, is applied to all the images $x_i$ in the dataset to extract the features from the last hidden layer (i.e., the value vectors returned by this layer). Using these features as new inputs and the labels $y_i$, we then train the classification layer for the task. After the training, we compute the Fisher information matrix for the feature extractor parameters. Since the Fisher information matrix is very large for deep networks, in practice we usually approximate it by (1) using only the diagonal entries and (2) averaging the Fisher information of all weights in the same filter. This results in a vector representation with size equal to the number of filters in the probe network. In this paper, we will use a ResNet~\cite{he2016deep} probe network that only has convolutional layers.

Task2Vec embeddings have many properties that can be used to study the relationships between tasks. We discuss two properties that are most relevant to our work. The first of these properties is that the norms of the embeddings encode the difficulty of the tasks. This property can be explained intuitively by noticing that easy examples (those that the model is very confident about) have less contributions to the Fisher information while uncertain examples (those that are near the decision boundary) have more contributions. Hence, if the task is difficult, the model would be uncertain on many examples leading to a large embedding.

The second property that we are interested in is that Task2Vec embeddings can encode the similarity between tasks. Achille et al.~\cite{achille2019task2vec} empirically showed this effect on the iNaturalist dataset~\cite{zhang2017imaterialist}, where the distances between Task2Vec embeddings strongly agree with the distances between natural taxonomical orders, hinting that the dissimilarity between tasks can be approximated from the distance between them in the embedding space. The embeddings were also shown to be useful for model selection between different domains and tasks.

\subsection{Total complexity}\label{sec:total}

We now discuss the notions of total complexity and sequential heterogeneity of task sequences, and how we can estimate them from Task2Vec embeddings. We note that these definitions only capture specific aspects of sequence complexity and heterogeneity; however, they are enough to serve the purpose of our paper. In future work, we will consider more sophisticated definitions of sequence complexity and heterogeneity.

We define the total complexity of a task sequence as the sum of the complexities of its individual tasks. Formally, let ${ T = (t_1, t_2, \ldots, t_k) }$ be a sequence of $k$ \emph{distinct} tasks and $C(t)$ be a function measuring the complexity of a task $t$. The total complexity of the task sequence $T$ is:
\begin{equation}
C(T) = \sum_{i=1}^k C(t_i).\label{eq:totalcomplex}
\end{equation}
We slightly abuse notation by using the same function $C(\cdot)$ for the complexity of both sequences and tasks.

For simplicity, we only consider sequences of distinct tasks where data for each task are only observed once. The scenario where data for one task may be observed many times requires different definitions of total complexity and sequential heterogeneity. We will leave this extension to future work.

A simple way to estimate the complexity $C(t)$ of a task $t$ is to measure the error rate of a model trained for this task. However, this method often gives unreliable estimates since it depends on various factors such as the choice of model and the training algorithm.

In this work, we propose to estimate $C(t)$ from the Task2Vec embedding of task $t$. Specifically, we adopt the suggestion from Achille et al. \cite{achille2019task2vec} to measure the complexity of task $t$ by its distance to the trivial task (i.e., the task embedded at the origin for standard Fisher embedding) in the embedding space. That is,
\begin{equation}
C(t) = d(e_t, e_0),
\end{equation}
where $e_t$ and $e_0$ are the embeddings of task $t$ and the trivial task respectively, and $d(\cdot, \cdot)$ is a symmetric distance between two tasks in the embedding space. Following Achille et al. \cite{achille2019task2vec}, we choose $d(\cdot, \cdot)$ to be the normalized cosine distance:
\begin{equation}
d(e_1, e_2) = \cos\left(\frac{e_1}{e_1 + e_2}, \frac{e_2}{e_1 + e_2}\right),\label{eq:dist}
\end{equation}
where $e_1$ and $e_2$ are two task embeddings and the division is element-wise.
This distance was shown to be well correlated with natural distances between tasks \cite{achille2019task2vec}.

The total complexity in Eq.~\eqref{eq:totalcomplex} depends on the sequence length. We can also consider the total complexity per task, $C(T)/k$, which does not depend on sequence length. In our analysis, however, we will only consider sequences of the same length. Hence, our results are not affected whether total complexity or total complexity per task is used.

We note that our total complexity measure is very crude and only captures some aspects of task sequence complexity. However, as we will show in Sec.~\ref{sec:results}, our measure is positively correlated with catastrophic forgetting and thus can be used to explain catastrophic forgetting. A possible future research direction would be to design better measures of task sequence complexity that can better explain catastrophic forgetting (i.e., by giving better correlation scores).

\subsection{Sequential heterogeneity}

We define the sequential heterogeneity of a task sequence as the sum of the dissimilarities between all pairs of consecutive tasks in the sequence. Formally, for a task sequence ${ T = (t_1, t_2, \ldots, t_k) }$ of distinct tasks, its sequential heterogeneity is:
\begin{equation}
F(T) = \sum_{i=1}^{k-1} F(t_i, t_{i+1}),\label{eq:seqhet}
\end{equation}
where $F(t, t')$ is a function measuring the dissimilarity between tasks $t$ and $t'$.
Note that we also use the same notation $F(\cdot)$ for sequential heterogeneity and task dissimilarity here, but its interpretation should be clear from the context.

The dissimilarity $F(t, t')$ can be naively estimated by applying transfer learning algorithms and measuring how well we can transfer between the two tasks. However, this would give a tautological measure of dissimilarity that is affected by both the model choice and the choice of the transfer learning algorithm.

To avoid this problem, we also propose to estimate $F(t, t')$ from the Task2Vec embedding. For our purpose, it is clear that we can use the distance $d(\cdot, \cdot)$ of Eq.~\eqref{eq:dist} as an estimate for $F(\cdot, \cdot)$. That is,
\begin{equation}
F(t, t') = d(e_t, e_{t'}) = \cos\left(\frac{e_t}{e_t + e_{t'}}, \frac{e_{t'}}{e_t + e_{t'}}\right).
\end{equation}

The sequential heterogeneity in Eq.~\eqref{eq:seqhet} only considers pairs of consecutive tasks, under the assumption that catastrophic forgetting is mostly influenced by the dissimilarity between these task pairs. In general, we can define other measures of heterogeneity, such as the total dissimilarity between all pairs of tasks. We will leave these extensions to future work.

Our choice of using Task2Vec to estimate $C(T)$ and $F(T)$ is more compatible with the multi-head models for continual learning~\cite{nguyen2018variational, zenke2017continual}, which we will use in our experiments. In multi-head models, a separate last layer (the SoftMax layer) is trained for each different task and the other weights are shared among tasks. This setting is consistent with the way Task2Vec is constructed in many cases. For instance, if we have two binary classification tasks whose labels are reversed, they would be considered similar by Task2Vec and are indeed very easy to transfer from one to another in the multi-head setting, by changing only the head.

\section{Correlation analysis}\label{sec:corr}

Having defined total complexity and sequential heterogeneity, we now discuss how we can study their relationships to the hardness of a task sequence. Given a task sequence $T = (t_1, t_2, \ldots, t_k)$, we measure its actual hardness with respect to a continual learning algorithm $A$ by the final error rate obtained after running $A$ on the tasks $t_1, t_2, \ldots, t_k$ sequentially. That is, the hardness of $T$ with respect to $A$ is:
\begin{equation}
H_A(T) = \mathrm{err}_A(T). \label{eq:hard}
\end{equation}

In this paper, we choose final error rate as the measure of actual hardness as it is an important metric commonly used to evaluate continual learning algorithms. In future work, we will explore other metrics such as the forgetting rate~\cite{chaudhry2018riemannian}.

To analyze the relationships between the hardness and total complexity or sequential heterogeneity, we employ correlation analysis as the main statistical tool. In particular, we sample $M$ task sequences $T_1, T_2, \ldots, T_M$ and compute their hardness measures $( H_A(T_i) )_{i=1}^M$ as well as their total complexity $( C(T_i) )_{i=1}^M$ and sequential heterogeneity $( F(T_i) )_{i=1}^M$ measures. From these measures, we compute the Pearson correlation coefficients between hardness and total complexity measures or between hardness and sequential heterogeneity measures. These coefficients tell us how correlated these quantities are.

Formally, the Pearson correlation coefficient between two corresponding sets of measures $X = ( X_i )_{i=1}^M$ and ${Y = ( Y_i )_{i=1}^M}$ is defined as:
\begin{equation}
r_{XY} = \frac{\sum_{i=1}^M (X_i - \bar{X}) (Y_i - \bar{Y})}{\sqrt{\sum_{i=1}^M (X_i - \bar{X})^2} \sqrt{\sum_{i=1}^M (Y_i - \bar{Y})^2}},
\end{equation}
where $\bar{X}$ and $\bar{Y}$ are the means of $X$ and $Y$ respectively. In addition to the correlation coefficients, we can also compute the p-values, which tell us how statistically significant these correlations are.

When computing the correlations between the hardness measures $H_A(T_i)$ and the total complexities $C(T_i)$, it is often a good idea to constrain the task sequences $T_i$ to have the same length. The reason for this normalization is that longer sequences tend to have larger complexities, thus the correlation may be biased by the sequence lengths rather than reflecting the complexity of individual tasks.

Similarly, when computing the correlations between $H_A(T_i)$ and the sequential heterogeneity $F(T_i)$, it is also a good idea to constrain the total complexity of the task sequences $T_i$ to be the same, so that the individual tasks' complexities would not affect the correlations. This can be achieve by using the same set of individual tasks for all the sequences (i.e., the sequences are permutations of each other). We call the sequential heterogeneity obtained from this method the \emph{normalized sequential heterogeneity}.

\section{Experiments}
\label{sec:exp}

We next describe the settings of our experiments and discuss our results. More detailed discussions on the implications of these results to continual learning and catastrophic forgetting research are provided in Sec.~\ref{sec:discuss}.

\subsection{Settings}\label{sec:settings}

\minisection{Datasets and task construction} We conduct our experiments on two datasets: MNIST and CIFAR-10, which are the most common datasets used to evaluate continual learning algorithms. For each of these sets, we construct a more general \emph{split} version as follows. First, we consider all pairs of different labels as a unit binary classification task, resulting in a total of 45 unit tasks. From these unit tasks, we then create 120 task sequences of length five by randomly drawing, for each sequence, five unit tasks without replacement. 

We also construct 120 \emph{split} task sequences which are permutations of a fixed task set containing five random unit tasks to compute the normalized sequential heterogeneity. For each unit task, we train its Task2Vec embedding using a ResNet18~\cite{he2016deep} probe network pre-trained on a combined dataset containing both MNIST and CIFAR-10.

\minisection{Algorithms and network architectures} We choose two recent continual learning algorithms to analyze in our experiments: synaptic intelligence (SI)~\cite{zenke2017continual} and variational continual learning (VCL)~\cite{nguyen2018variational}. For the experiments on MNIST, we also consider the coreset version of VCL (coreset VCL). These algorithms are among the state-of-the-art continual learning algorithms on the considered datasets, with SI representing the regularization-based methods, VCL representing the Bayesian methods, and coreset VCL combining Bayesian and rehearsal methods.

On CIFAR-10, we run SI with the same network architecture as those considered in \cite{zenke2017continual}: a CNN with four convolutional layers, followed by two dense layers with dropout. Since VCL was not originally developed with convolutional layers, we flatten the input images and train with a fully connected network containing four hidden layers, each of which has 256 hidden units. On MNIST, we run both SI and VCL with a fully connected network containing two hidden layers, each of which has 256 hidden units. We denote this setting by MNIST-$256^2$.

Since MNIST is a relatively easy dataset, we may not observe meaningful results if all the errors obtained from different sequences are low and not very different. Thus, to make the dataset harder for the learning algorithms, we also consider smaller network architectures. In particular, we consider fully connected networks with a {\em single} hidden layer, containing either 50 hidden units (for MNIST-$50$) or 20 hidden units (for MNIST-$20$). Following \cite{nguyen2018variational, zenke2017continual}, we also use the multi-head version of the models where a separate last layer (the SoftMax layer) is trained for each different task and the other weights are shared among tasks. For coreset VCL, we use random coresets with sizes 40, 40, 20 for MNIST-$256^2$, MNIST-$50$ and MNIST-$20$ respectively.

\minisection{Optimizer settings} For both SI and VCL, we set the regularization strength parameter to the default value $\lambda=1$. In all of our experiments, the models are trained using Adam optimizer~\cite{kingma2014adam} with learning rate 0.001. Similar to \cite{zenke2017continual}, we set the batch size to be 256 in CIFAR-10 and 64 in MNIST settings for SI. We run this algorithm for 60, 10, 10, 5 epochs per task on CIFAR-10, MNIST-$256^2$, MNIST-$50$ and MNIST-$20$ respectively. 

For VCL and coreset VCL, we set the batch size to be the training set size~\cite{nguyen2018variational} and run the algorithms for 50, 120, 50, 20 epochs per task on CIFAR-10, MNIST-$256^2$, MNIST-$50$ and MNIST-$20$ respectively. For all algorithms, we run each setting ten times using different random seeds and average their errors to get the final error rates.

\subsection{Results}\label{sec:results}

\begin{figure*}[!t]
\begin{center}
\vskip -0.3in
\begin{tabular}{cccc}
\raisebox{0.5in}{\rotatebox[origin=t]{90}{SI}} &
\includegraphics[width=0.25\textwidth]{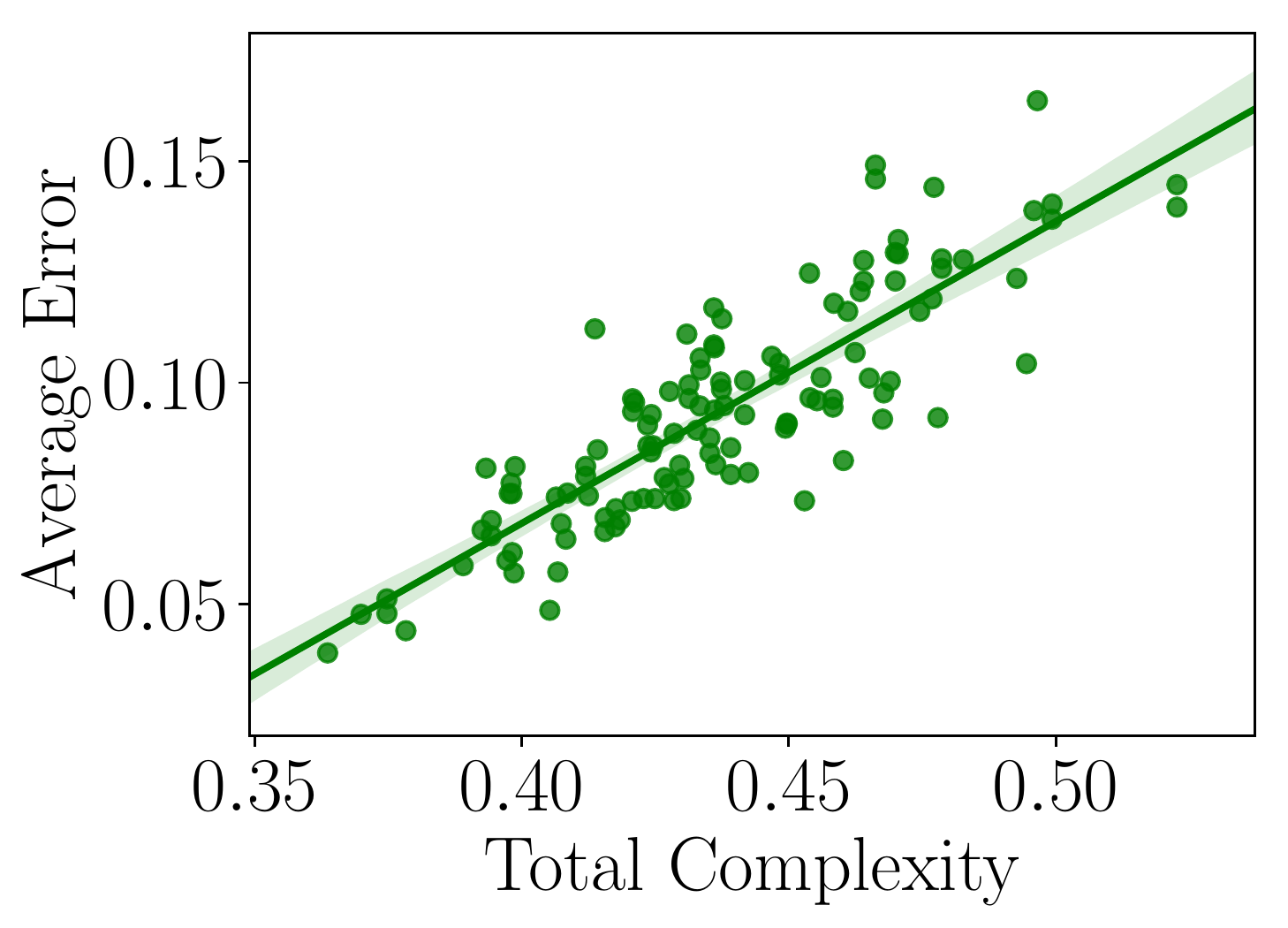} \hspace{0.5cm} &
\includegraphics[width=0.25\textwidth]{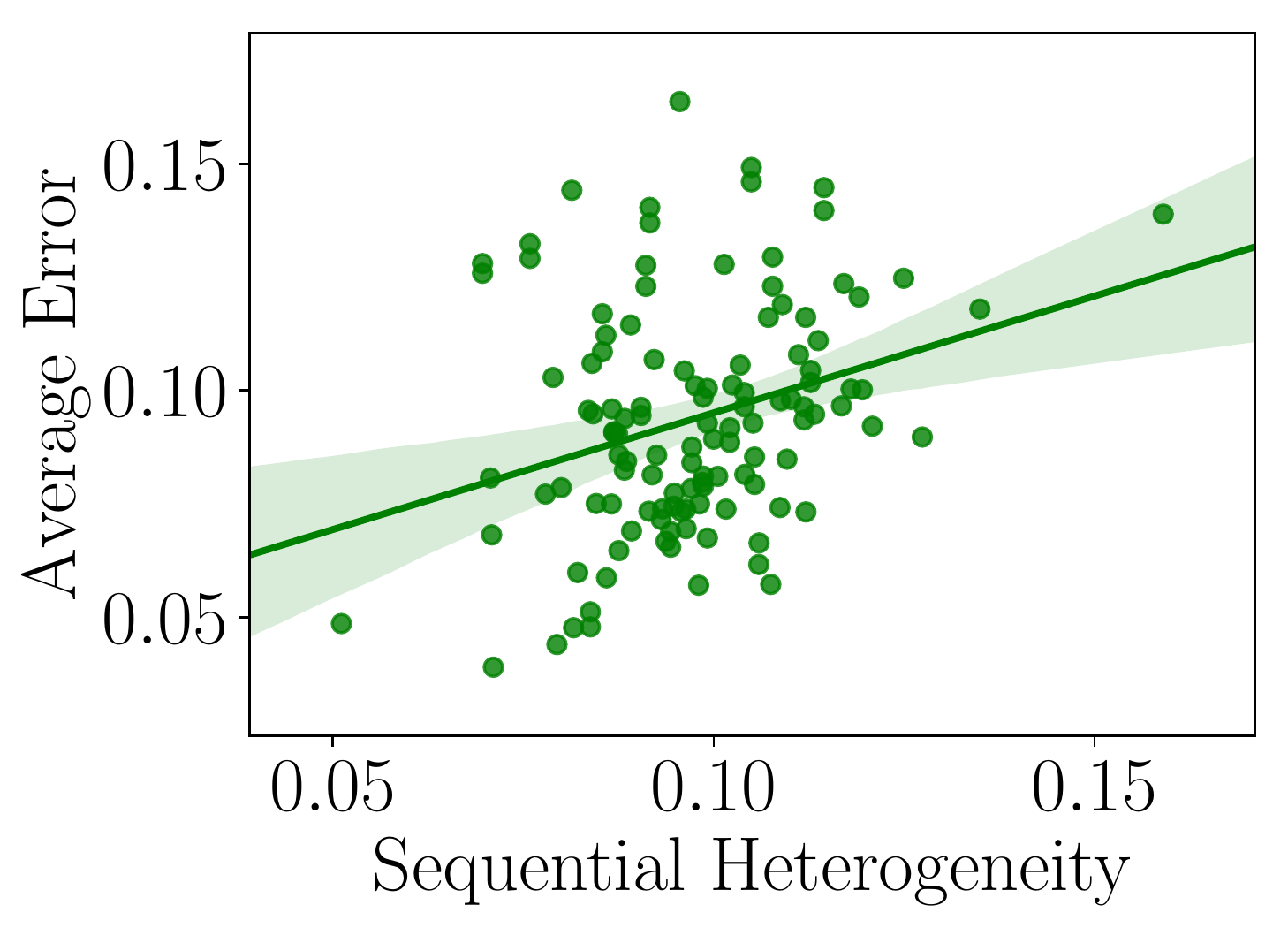} &
\includegraphics[width=0.25\textwidth]{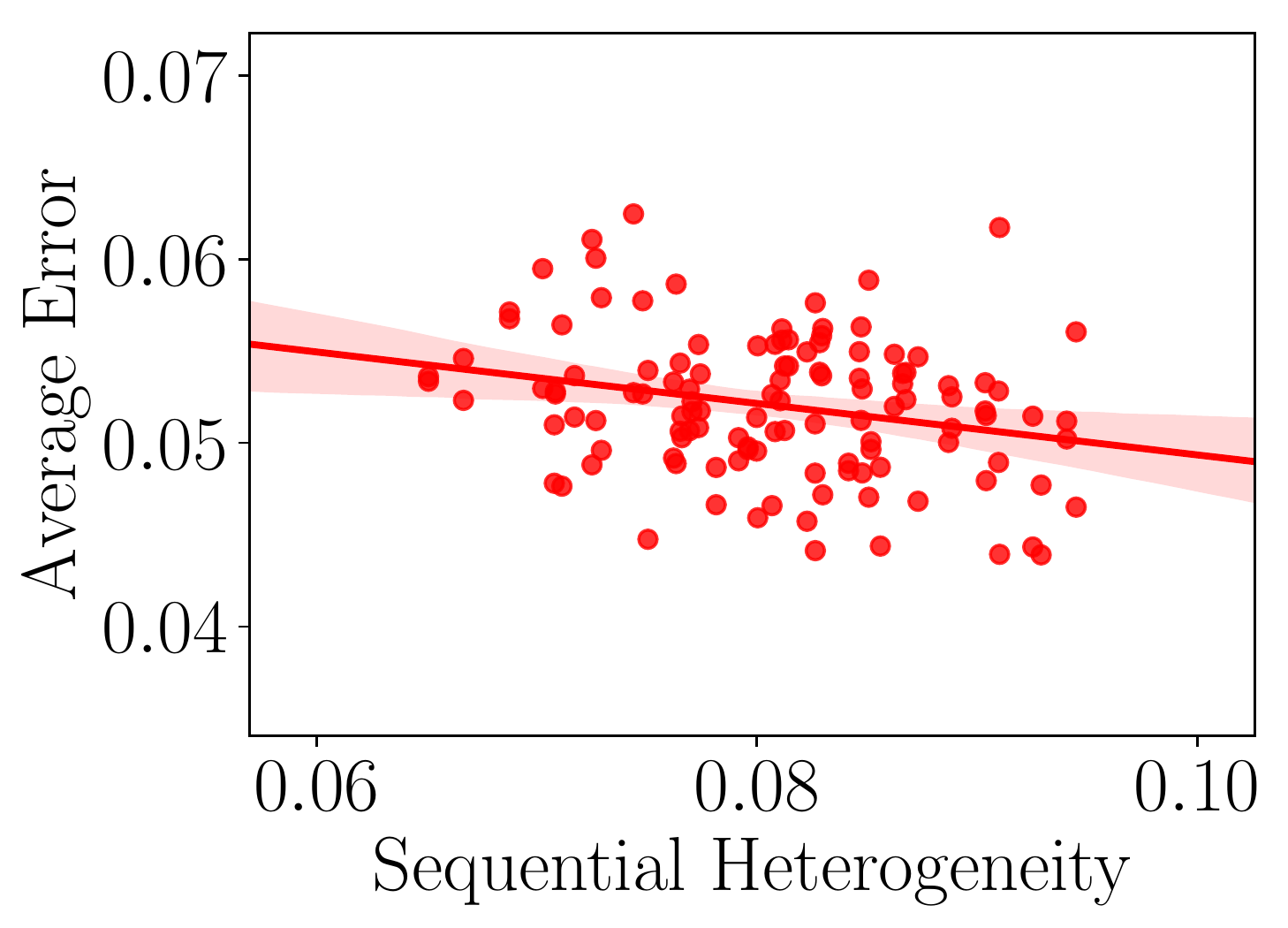} \\
\raisebox{0.5in}{\rotatebox[origin=t]{90}{VCL}} &
\includegraphics[width=0.25\textwidth]{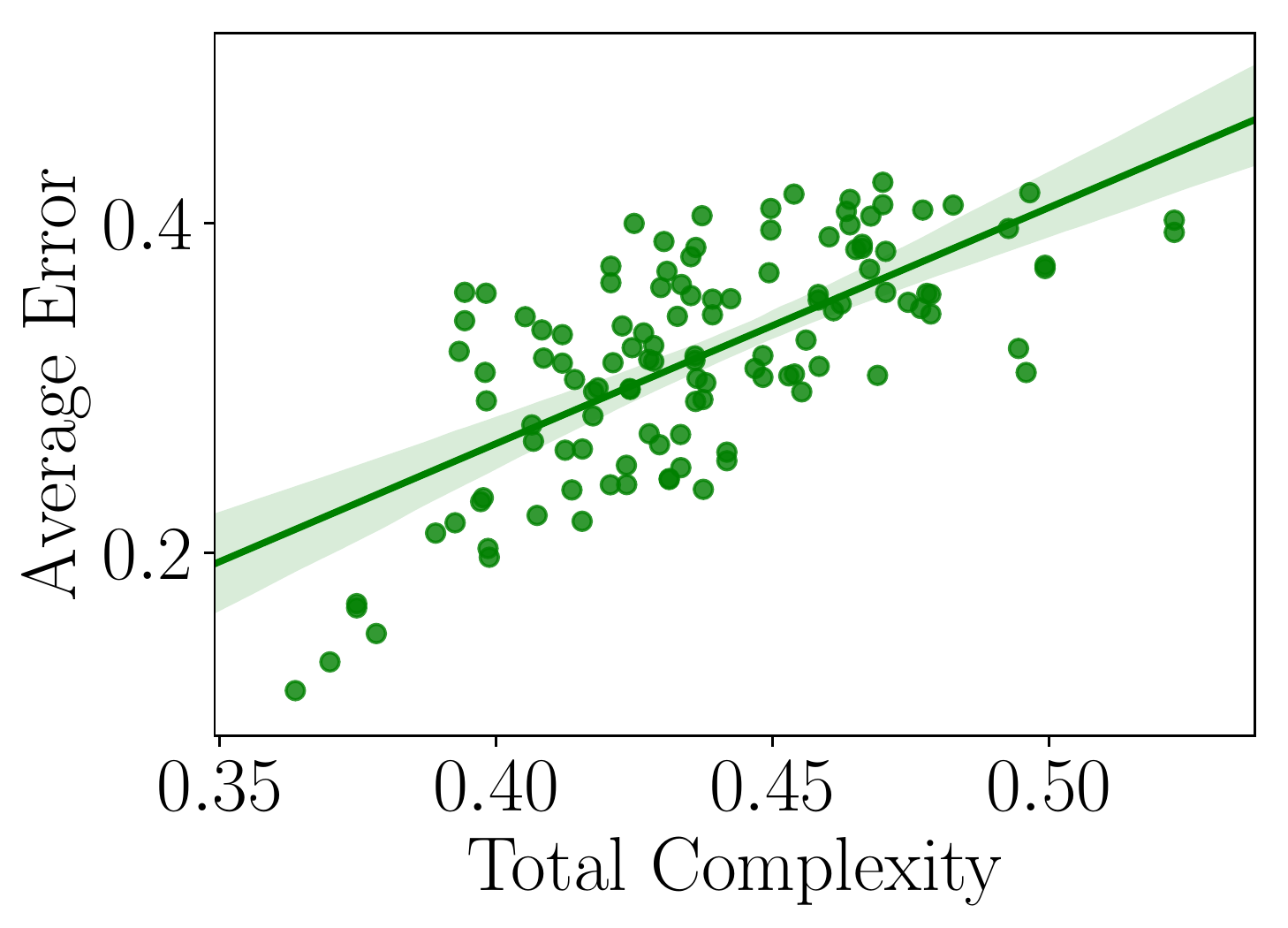} \hspace{0.5cm} &
\includegraphics[width=0.25\textwidth]{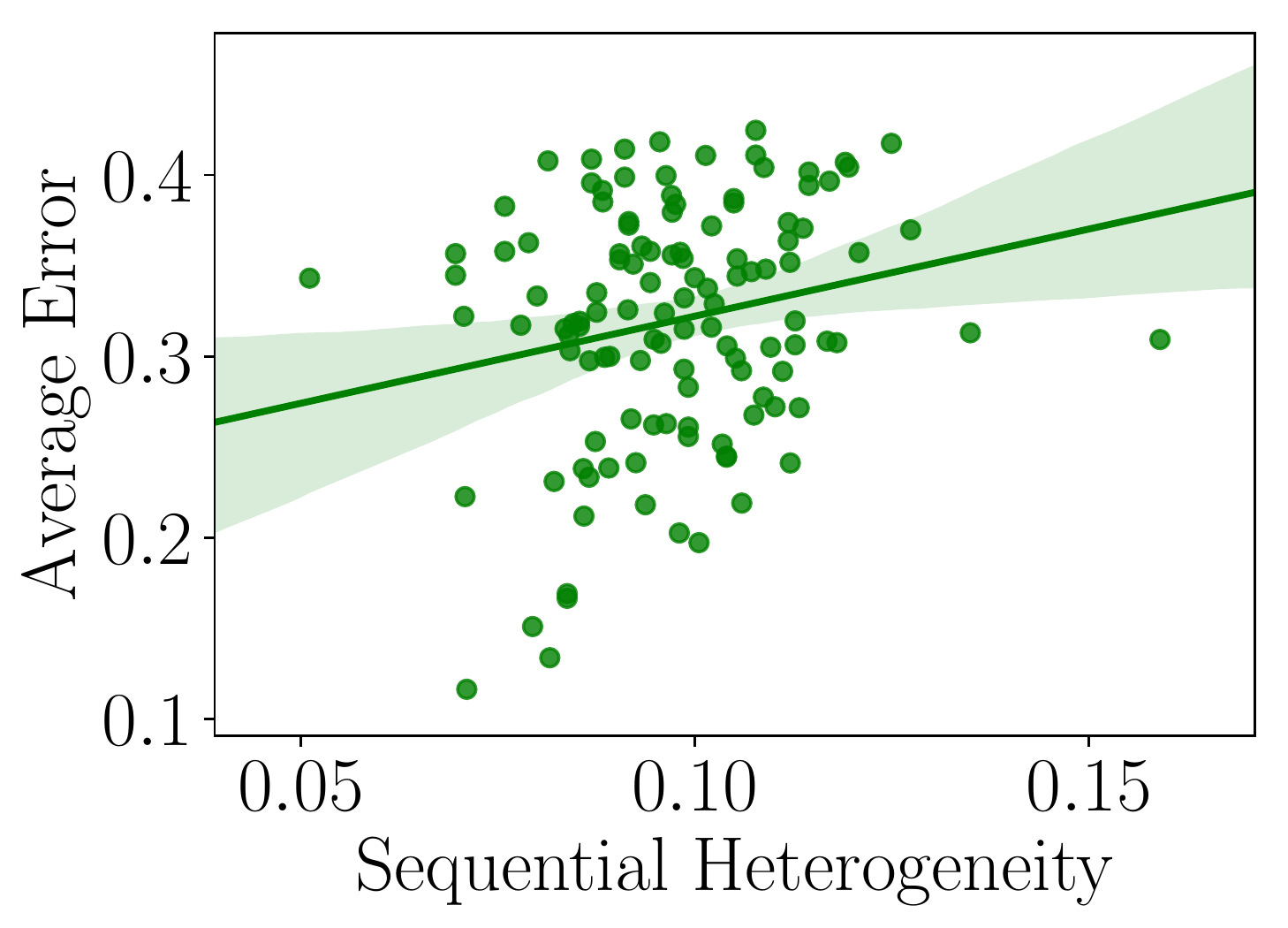} &
\includegraphics[width=0.25\textwidth]{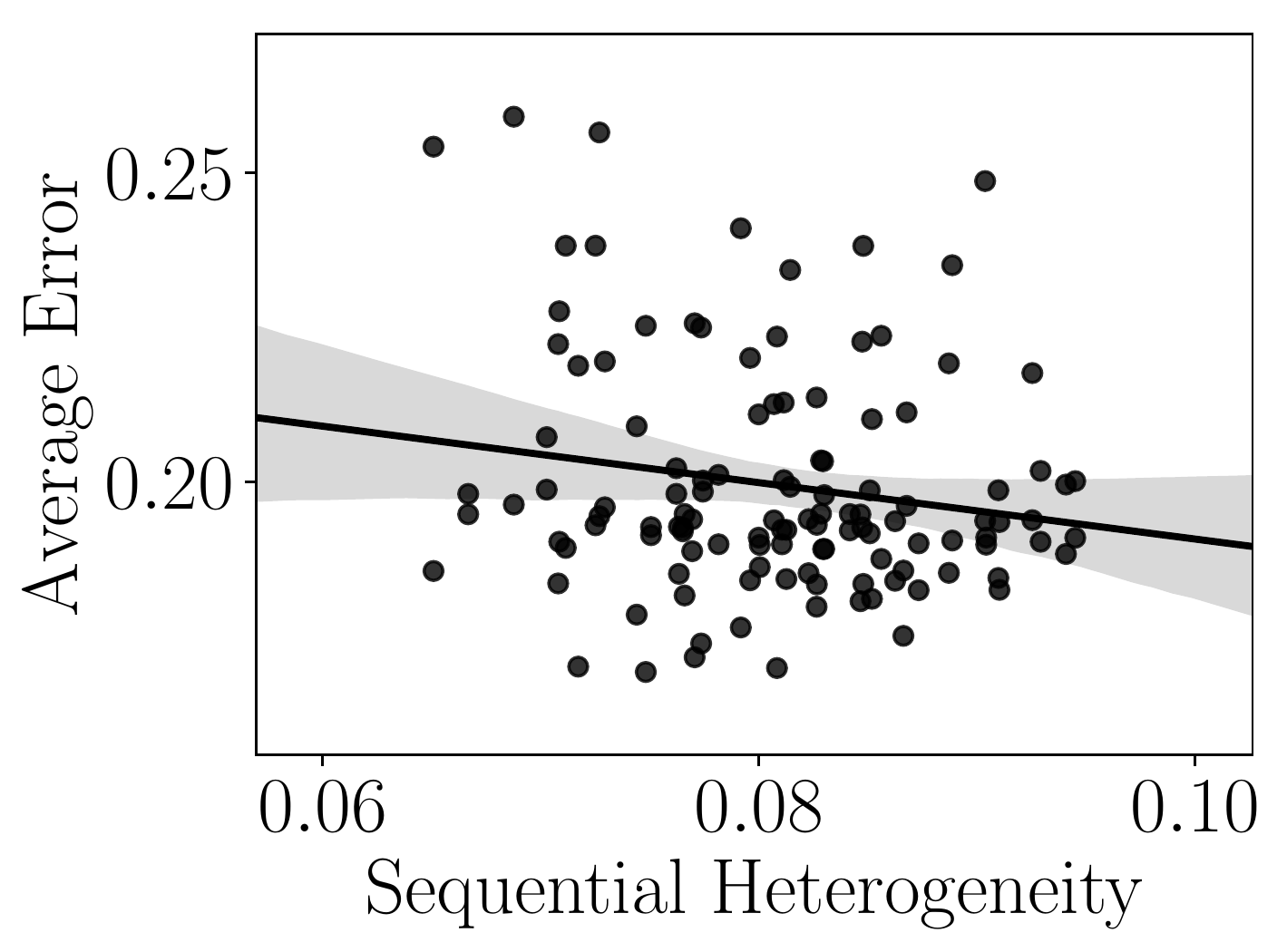} \\
& (a) Total complexity & (b) Sequential heterogeneity  & (c) Normalized sequential heterogeneity
\end{tabular}
\end{center}
\vskip -0.1cm
\caption{{\bf Error vs. (a) total complexity, (b) sequential heterogeneity and (c) normalized sequential heterogeneity on CIFAR-10}, together with the linear regression fits and 95\% confidence intervals. Green (red) color indicates statistically significant positive (negative) correlations. Black color indicates negligible correlations.}
\label{fig:cifar}
\vskip 0.2cm
\end{figure*}

\begin{table*}[!ht]
\begin{center}
\begin{tabular}{ccccccc}
\toprule
& Variable & Algorithm & MNIST-$256^2$ & MNIST-$50$ & MNIST-$20$ & CIFAR-10 \\
\midrule
\textnormal{(a)} & Total & SI & \bf{0.24} ($p<0.01$) & \bf{0.22} ($p<0.05$) & \bf{0.36} ($p<0.01$) & \bf{0.86} ($p<0.01$) \\
& Complexity & VCL & 0.05 ($p=0.59$) & 0.17 ($p=0.07$) & \bf{0.21} ($p<0.05$) & \bf{0.69} ($p<0.01$) \\
& & Coreset VCL & \bf{0.28} ($p<0.01$) & \bf{0.41} ($p<0.01$) & \bf{0.37} ($p<0.01$) & - \\
\midrule
\textnormal{(b)} & Sequential & SI & -0.01 ($p=0.86$) & 0.05 ($p=0.55$) & 0.07 ($p=0.48$) & \bf{0.30} ($p<0.01$) \\
& Heterogeneity & VCL & 0.04 ($p=0.69$) & 0.01 ($p=0.88$) & 0.05 ($p=0.58$) & \bf{0.21} ($p<0.05$) \\
& & Coreset VCL & 0.09 ($p=0.31$) & 0.12 ($p=0.18$) & 0.18 ($p=0.05$) & - \\
\midrule
\textnormal{(c)} & Normalized & SI & -0.07 ($p=0.43$) & -0.04 ($p=0.65$) & 0.05 ($p=0.58$) & \bf{-0.25} ($p<0.01$) \\
& Sequential & VCL & 0.03 ($p=0.76$) & \bf{-0.20} ($p<0.05$) & \bf{-0.21} ($p<0.05$) & -0.17 ($p=0.06$) \\
& Heterogeneity & Coreset VCL & -0.08 ($p=0.37$) & \bf{-0.26} ($p<0.01$) & -0.16 ($p=0.07$) & - \\
\bottomrule
\end{tabular}
\end{center}
\caption{{\bf Correlation coefficients (p-values) between error rate and (a) total complexity, (b) sequential heterogeneity, and (c) normalized sequential heterogeneity} of three state-of-the-art continual learning algorithms (SI, VCL, coreset VCL) on four different tests conducted with the CIFAR-10 and MNIST datasets. Results with statistical significance ($p<0.05$) are shown in bold.}
\label{tab:result}
\end{table*}

\begin{figure*}[!h]
\begin{center}
\raisebox{0.5in}{\rotatebox[origin=t]{90}{SI}}
\hskip 0.1in
\stackon[2pt]{\includegraphics[width=0.22\textwidth]{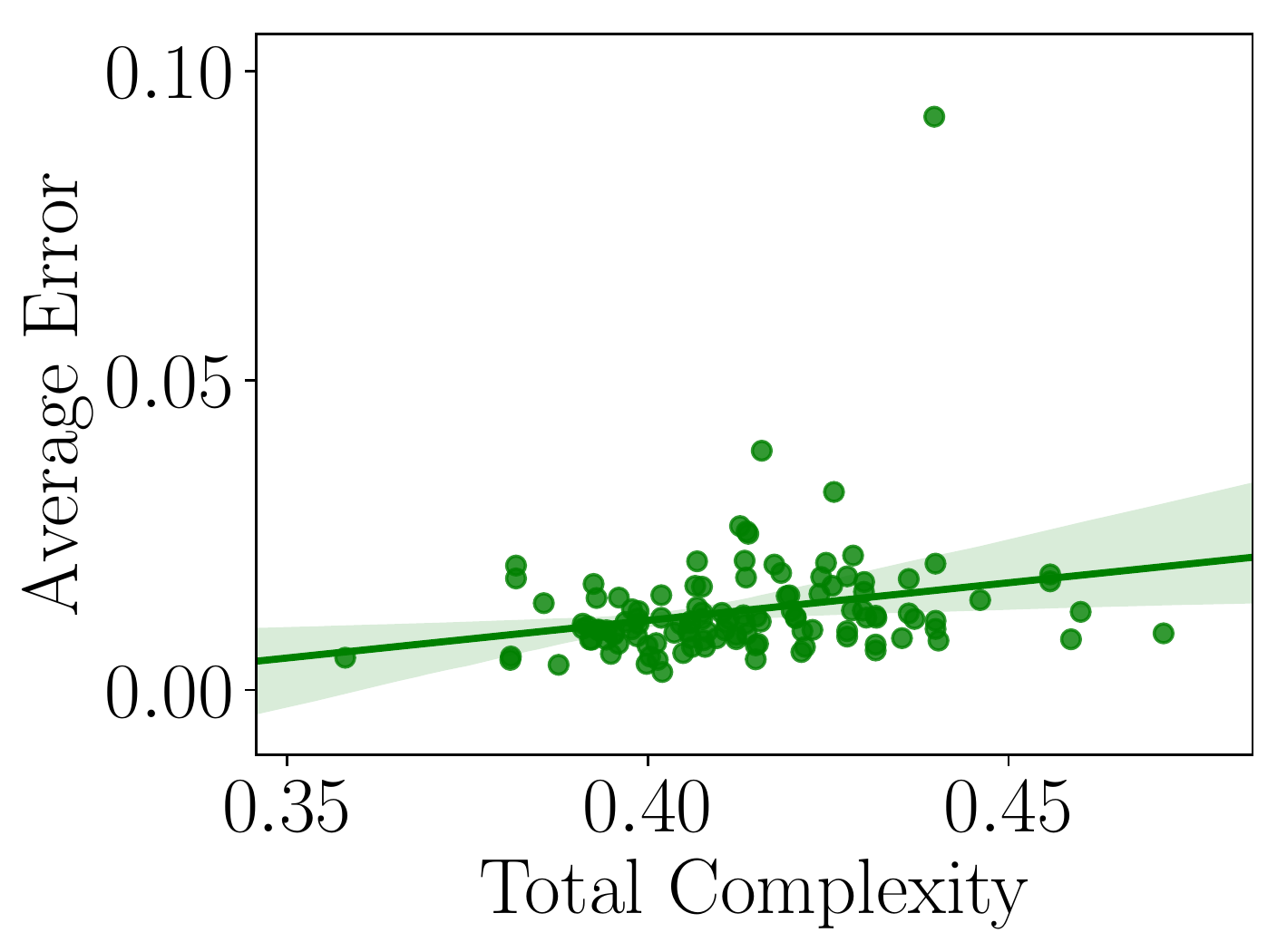}}{\quad MNIST-$256^2$}
\hskip 0.1in
\stackon[2pt]{\includegraphics[width=0.22\textwidth]{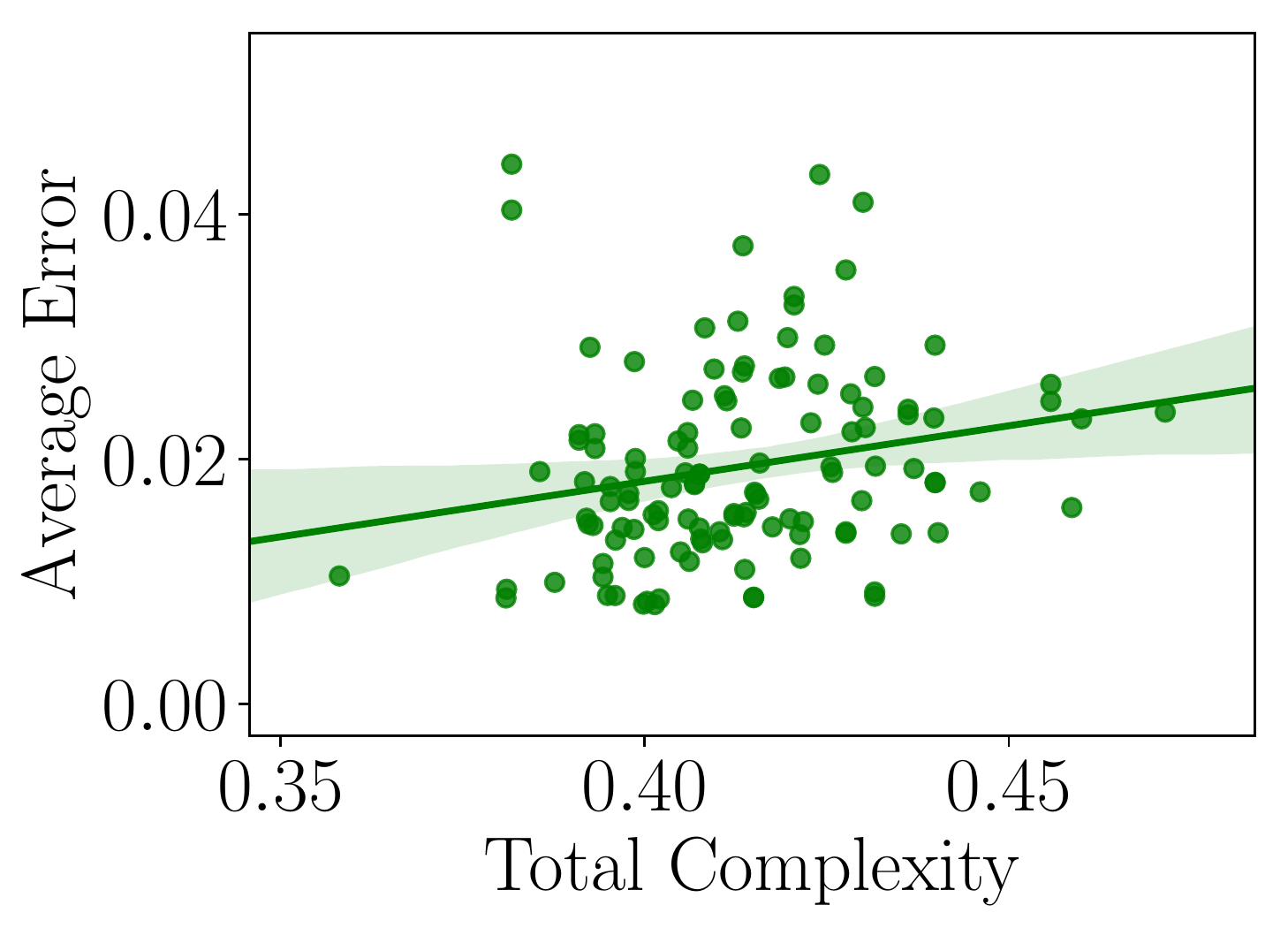}}{\quad MNIST-$50$}
\hskip 0.1in
\stackon[2pt]{\includegraphics[width=0.22\textwidth]{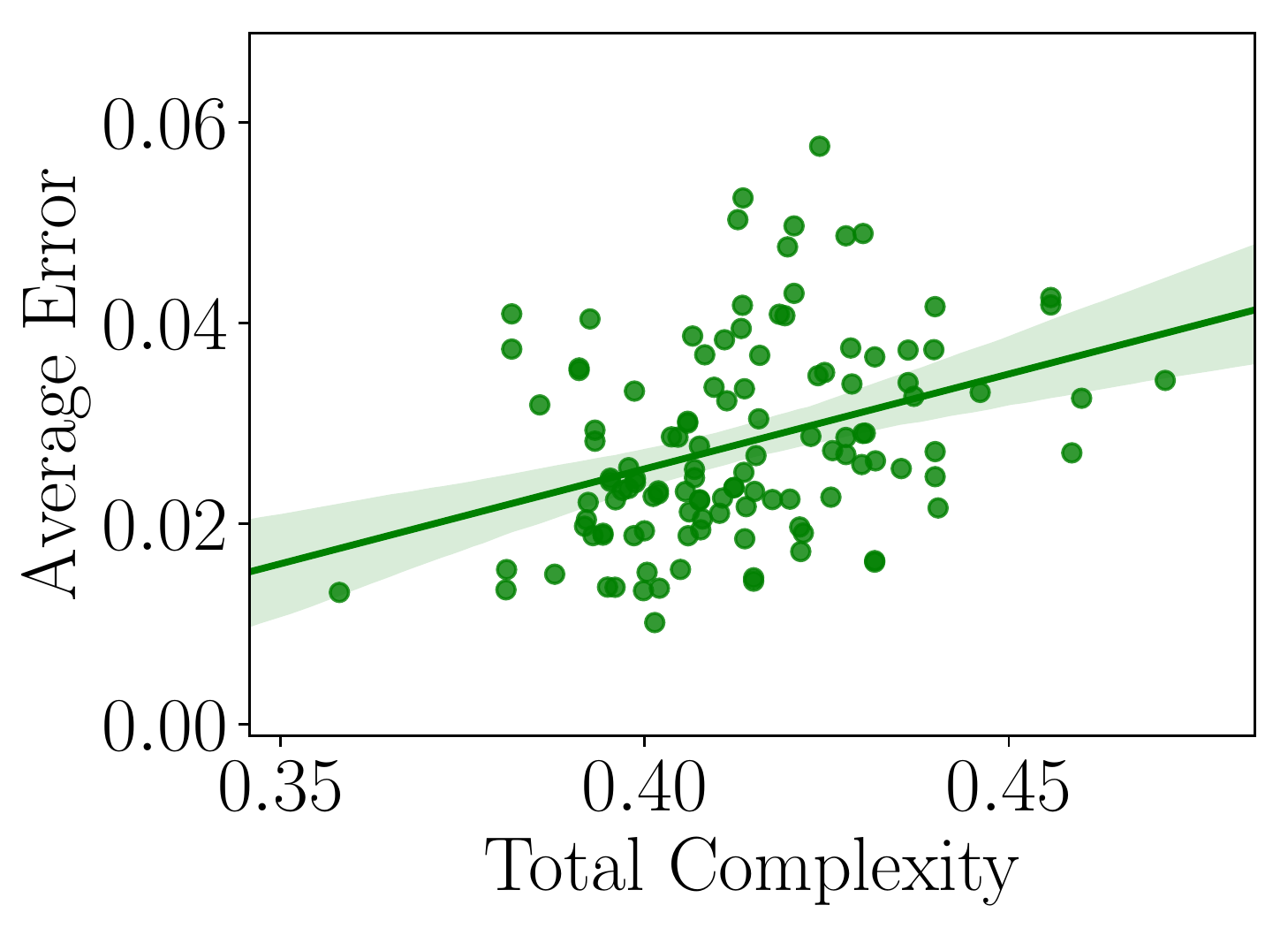}}{\quad MNIST-$20$}
\hskip 0.1in
\stackon[2pt]{\includegraphics[width=0.22\textwidth]{figs/com_cifar_si.pdf}}{\quad CIFAR-10}
\\
\raisebox{0.5in}{\rotatebox[origin=t]{90}{VCL}}
\hskip 0.1in
\includegraphics[width=0.22\textwidth]{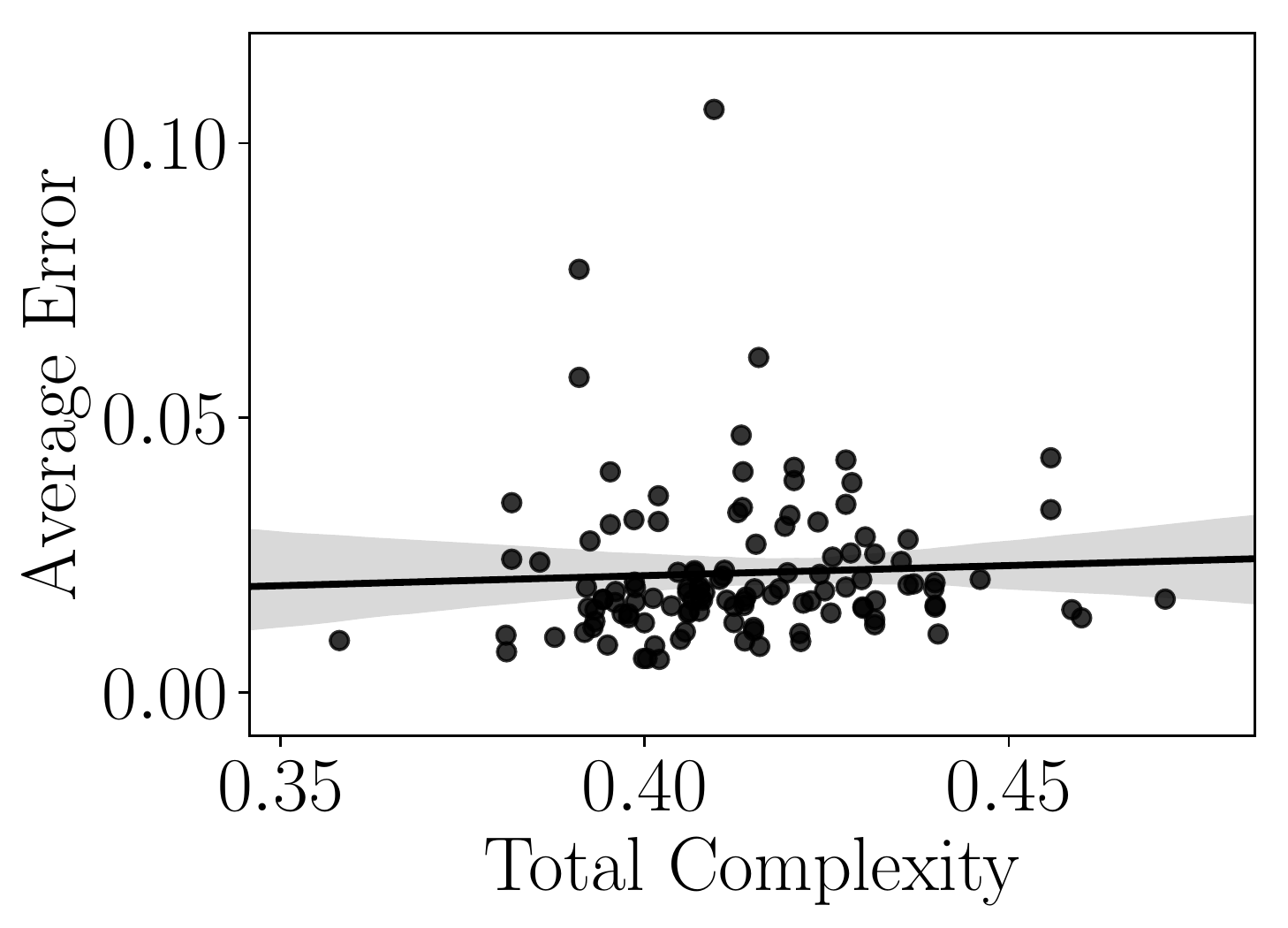}
\hskip 0.1in
\includegraphics[width=0.22\textwidth]{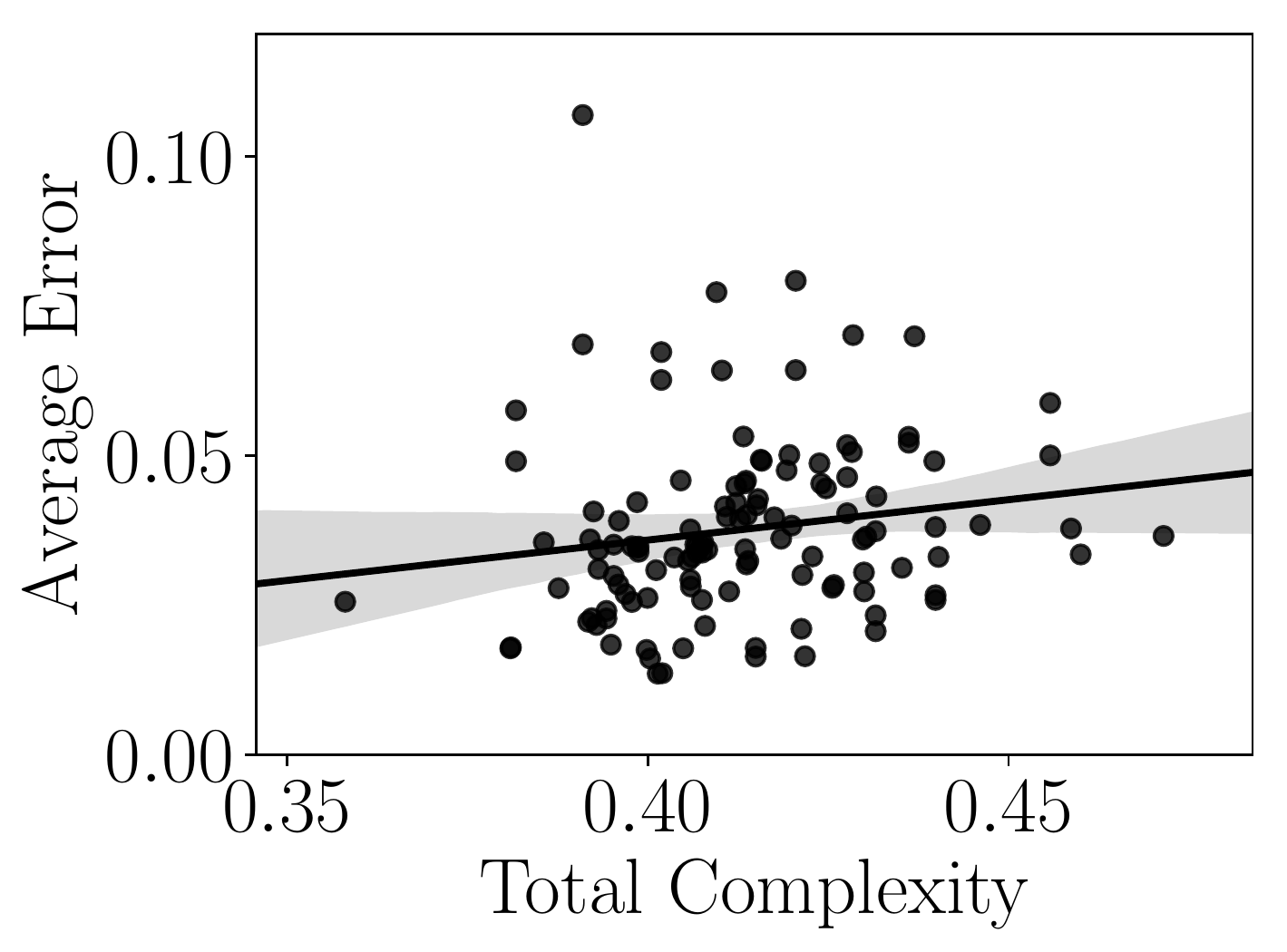}
\hskip 0.1in
\includegraphics[width=0.22\textwidth]{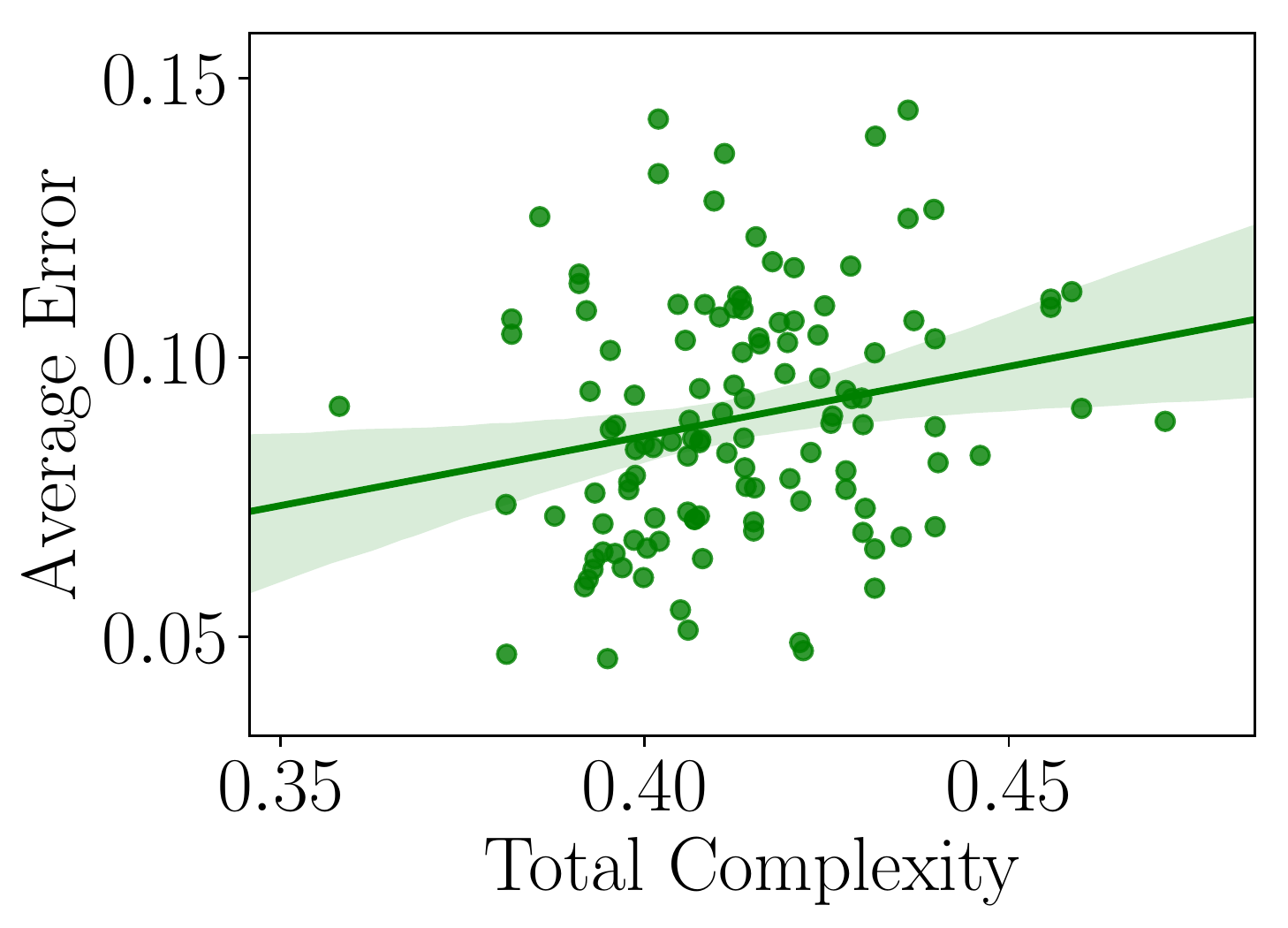}
\hskip 0.1in
\includegraphics[width=0.22\textwidth]{figs/com_cifar_vcl.pdf}
\\
\hskip -0.24\textwidth
\raisebox{0.5in}{\rotatebox[origin=t]{90}{coreset VCL}}
\hskip 0.1in
\includegraphics[width=0.22\textwidth]{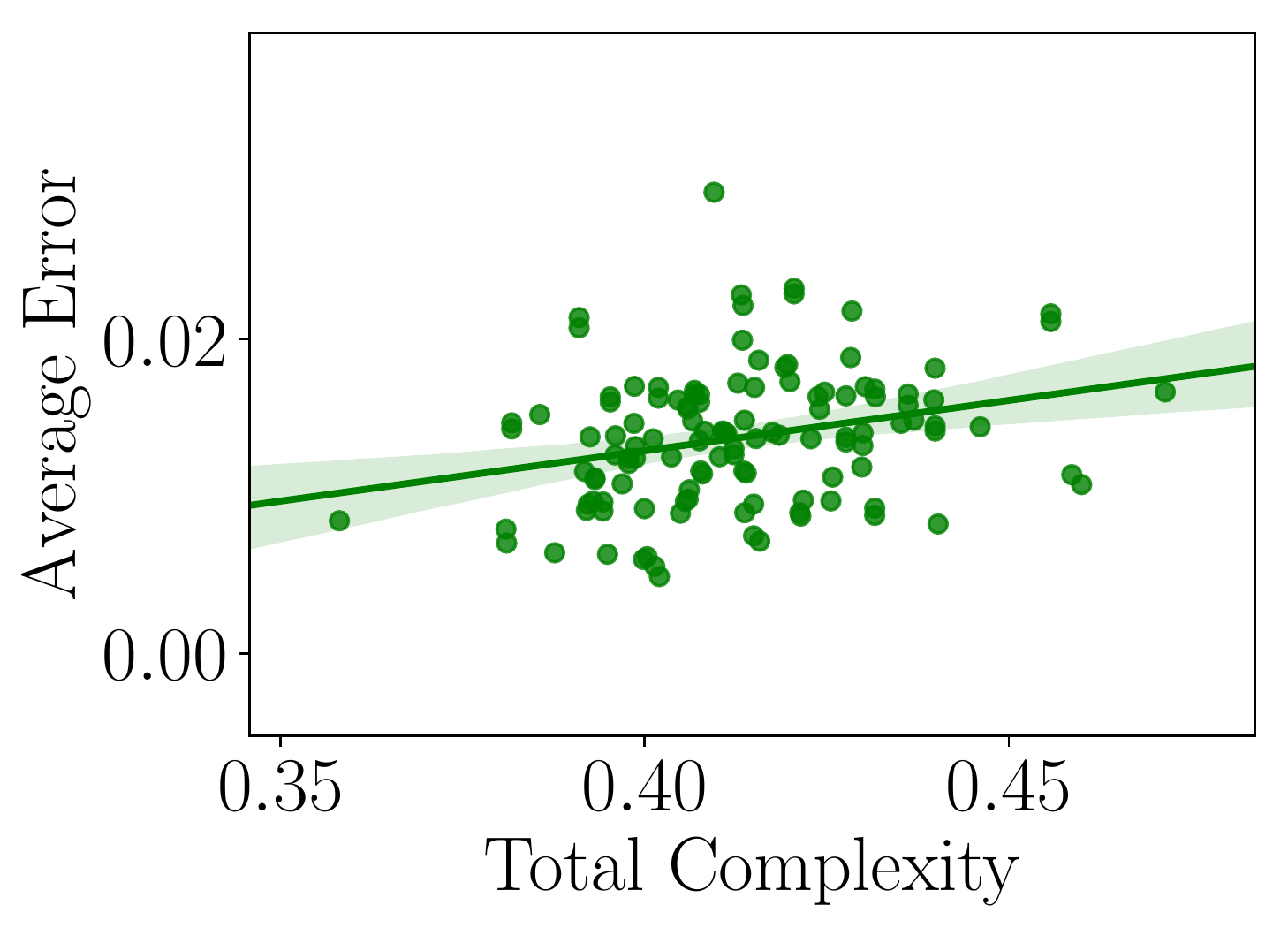}
\hskip 0.1in
\includegraphics[width=0.22\textwidth]{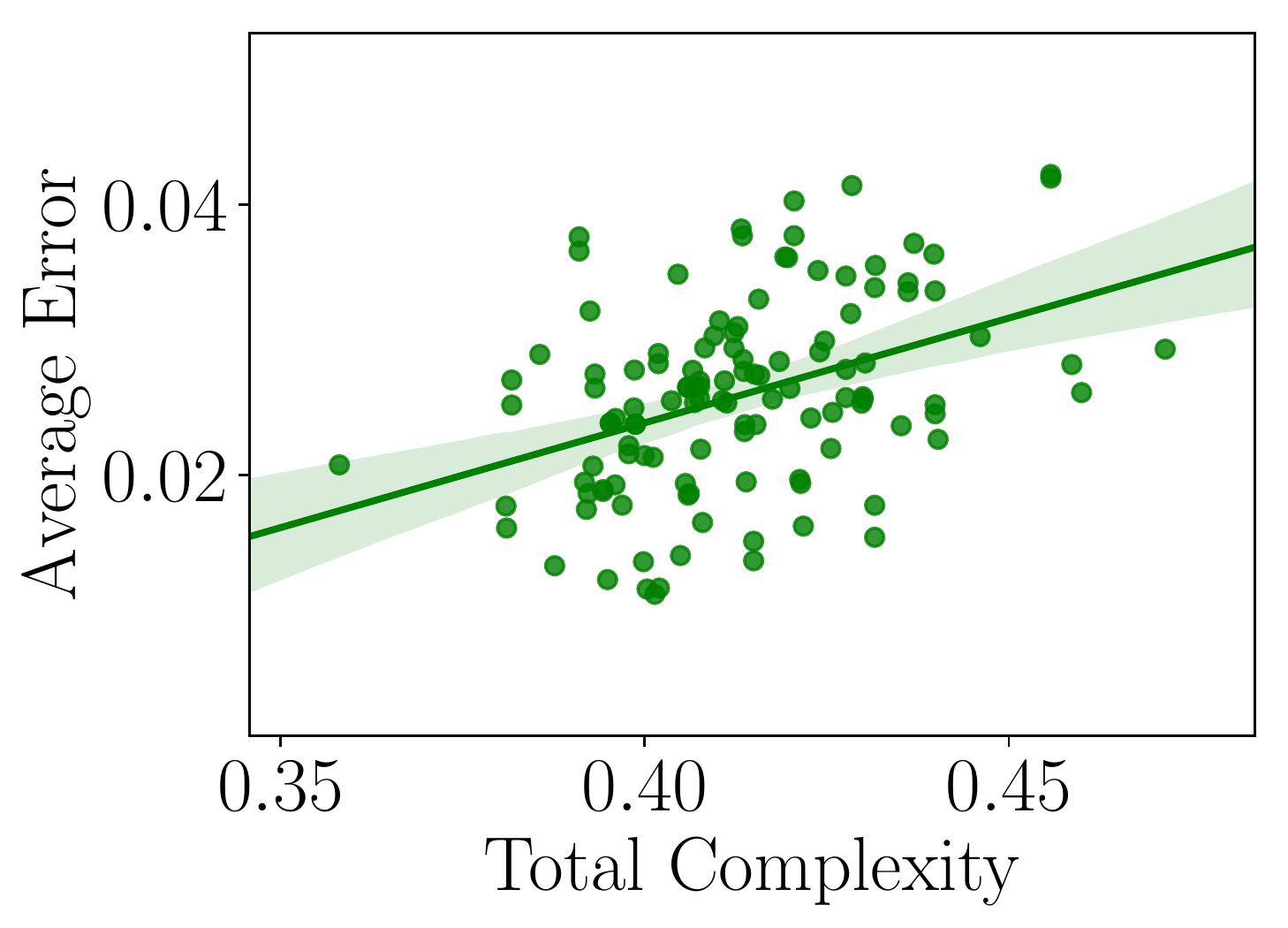}
\hskip 0.1in
\includegraphics[width=0.22\textwidth]{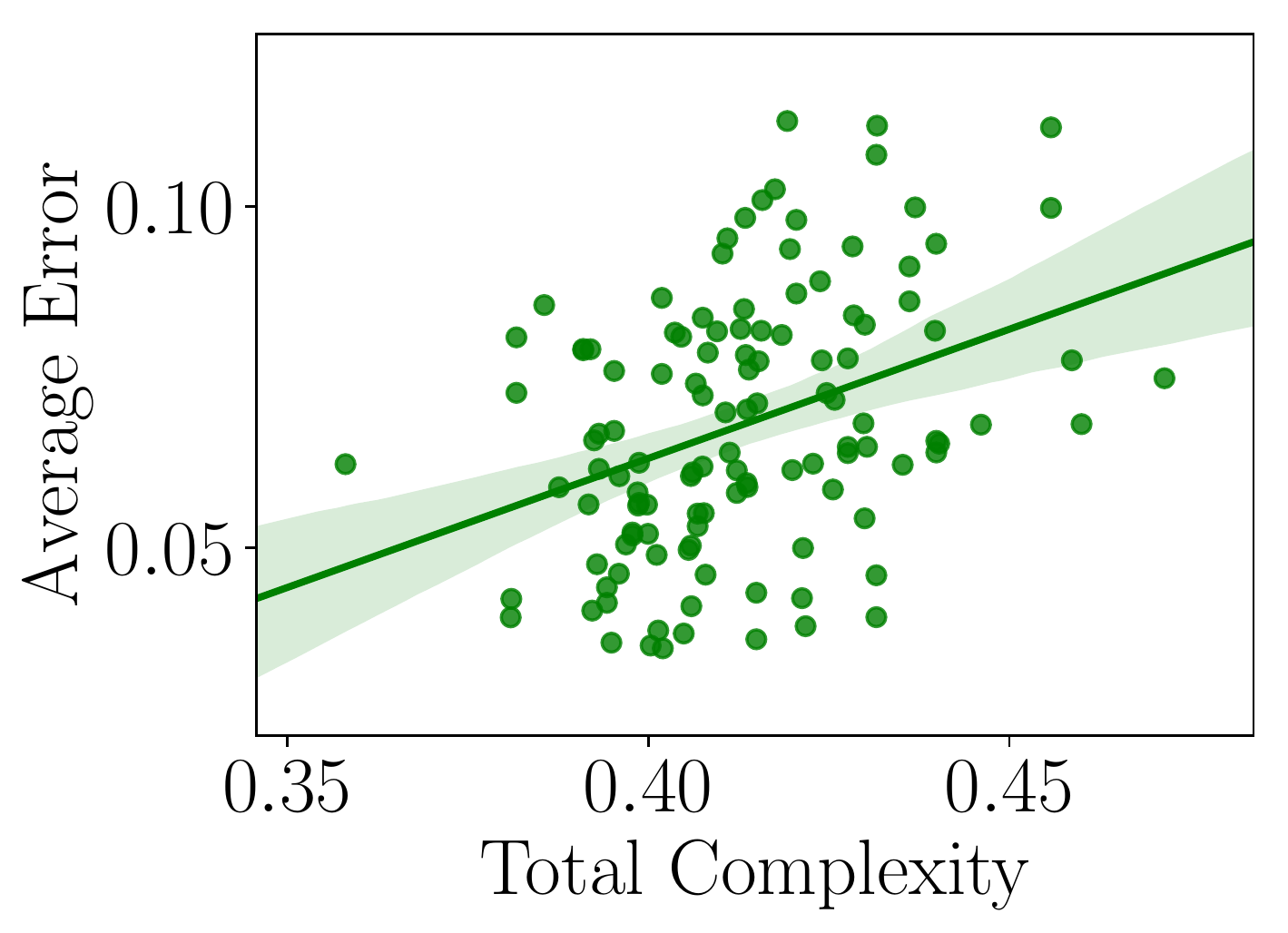}
\vskip 0.1in
\caption{{\bf Total complexity vs. average error}, together with the linear regression fit and 95\% confidence interval, for each algorithm and test in Table 1(a). Green color indicates statistically significant positive correlations. Black color indicates negligible correlations.}
\label{fig:complexity}

\vskip 0.5in

\raisebox{0.5in}{\rotatebox[origin=t]{90}{SI}}
\hskip 0.1in
\stackon[2pt]{\includegraphics[width=0.22\textwidth]{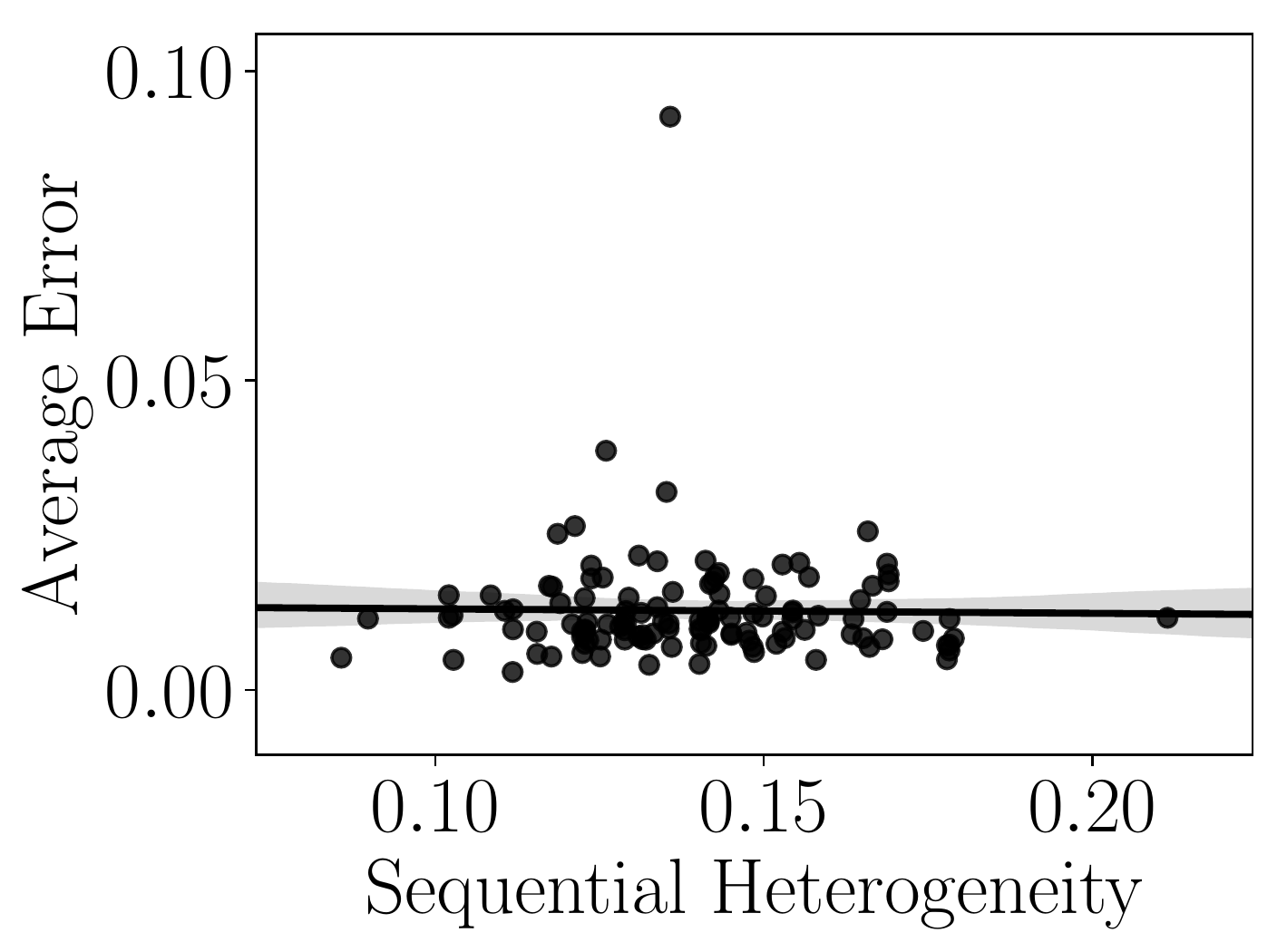}}{\quad MNIST-$256^2$}
\hskip 0.1in
\stackon[2pt]{\includegraphics[width=0.22\textwidth]{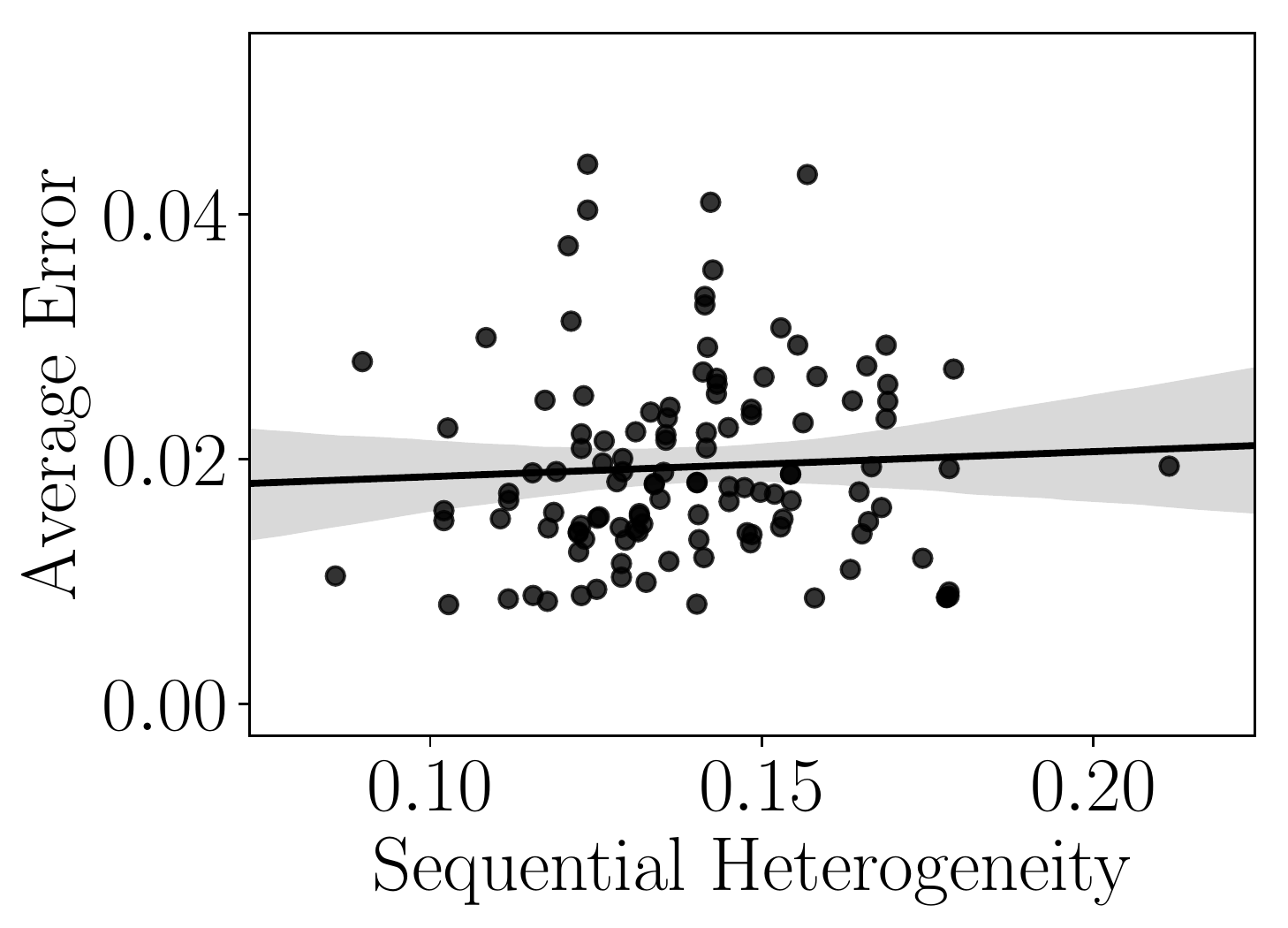}}{\quad MNIST-$50$}
\hskip 0.1in
\stackon[2pt]{\includegraphics[width=0.22\textwidth]{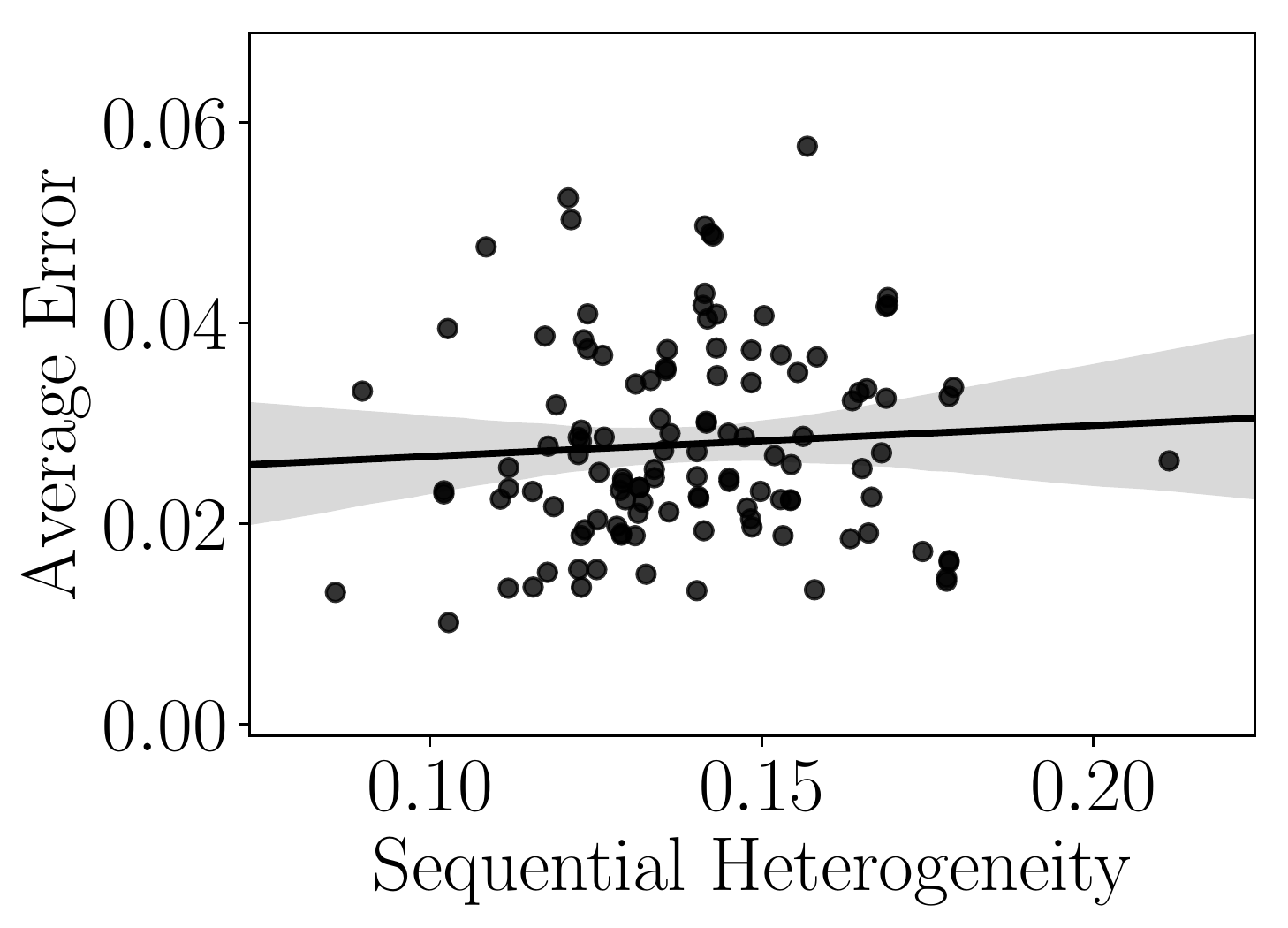}}{\quad MNIST-$20$}
\hskip 0.1in
\stackon[2pt]{\includegraphics[width=0.22\textwidth]{figs/hete_cifar_si.pdf}}{\quad CIFAR-10}
\\
\raisebox{0.5in}{\rotatebox[origin=t]{90}{VCL}}
\hskip 0.1in
\includegraphics[width=0.22\textwidth]{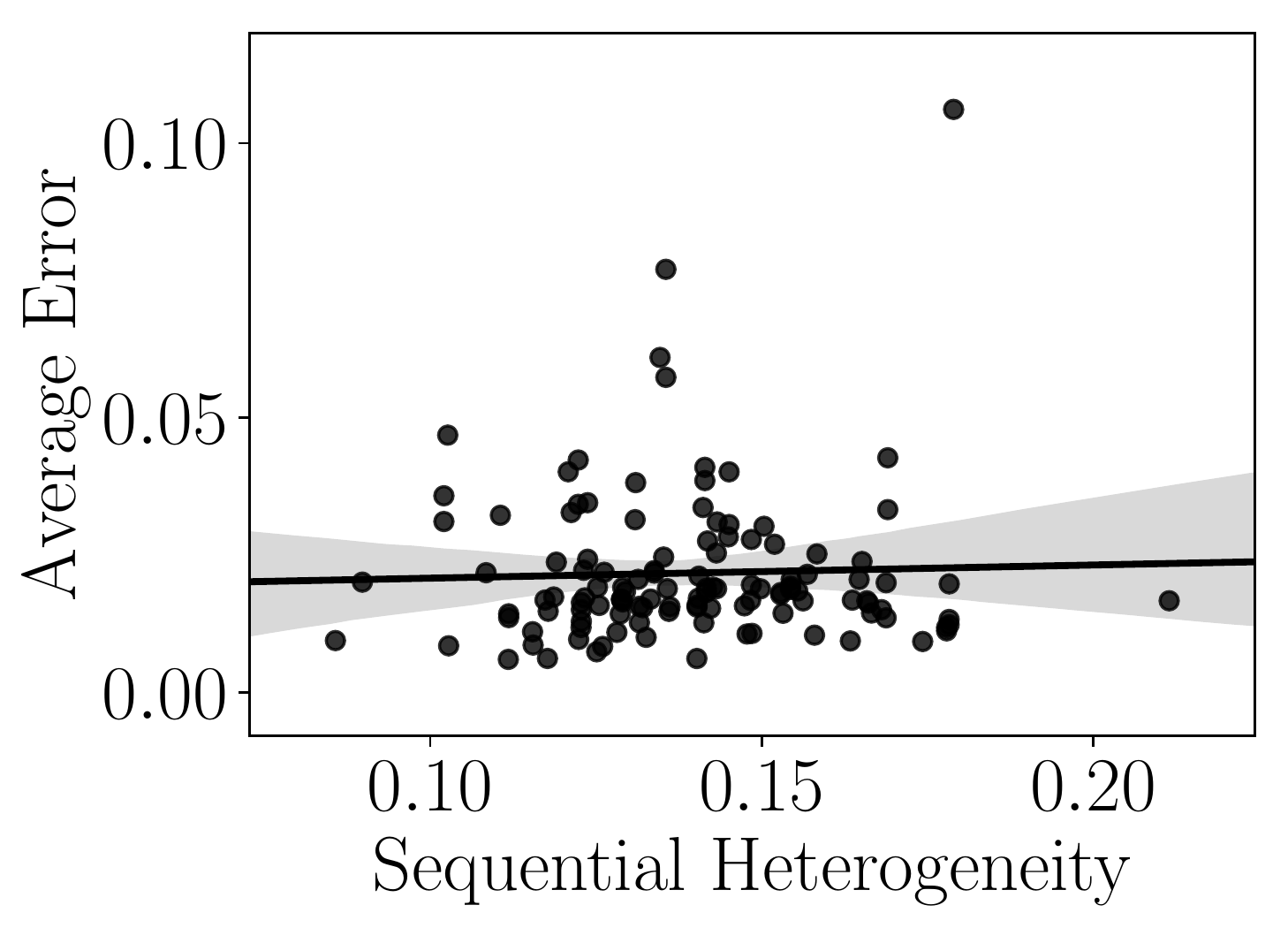}
\hskip 0.1in
\includegraphics[width=0.22\textwidth]{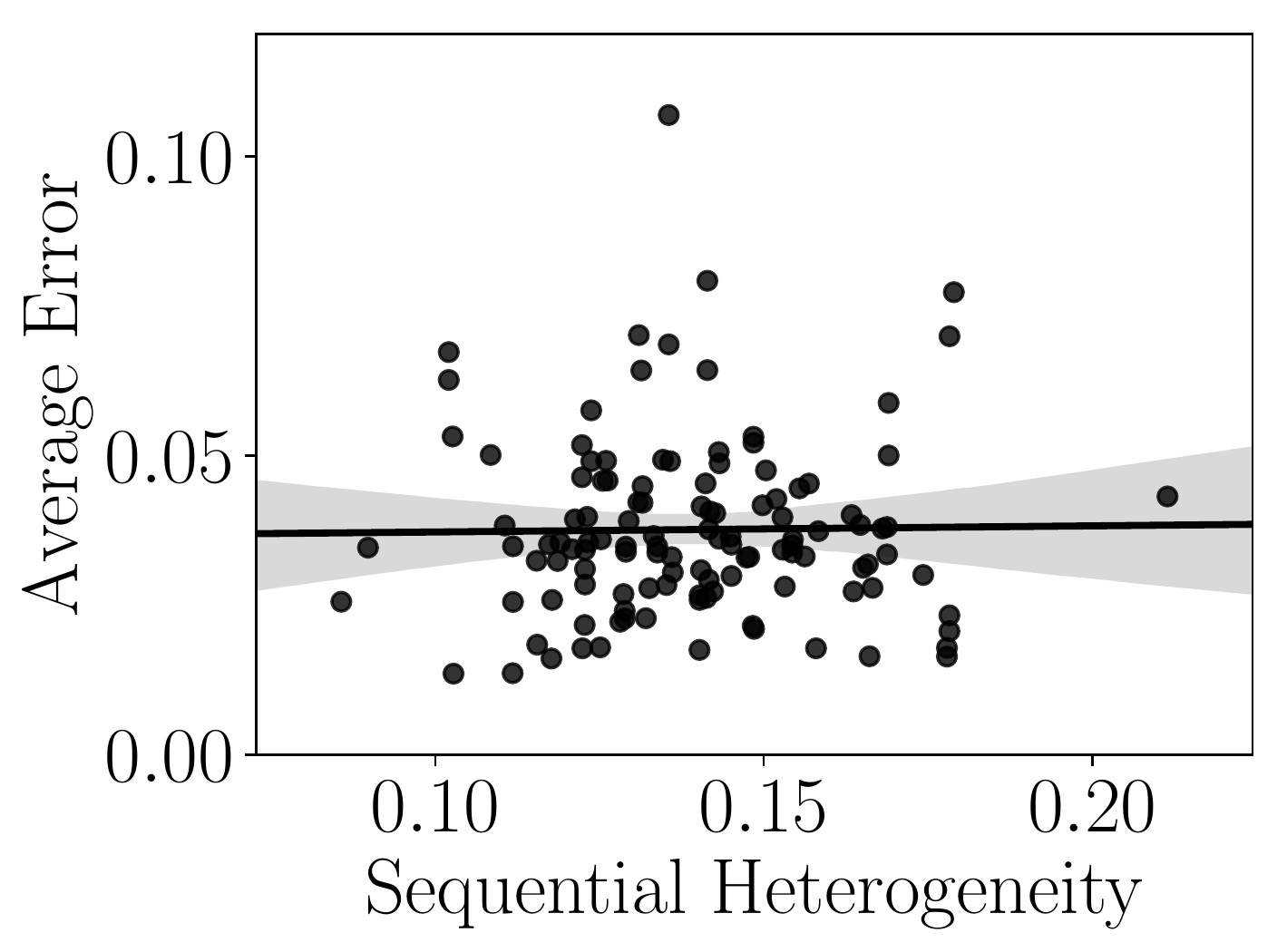}
\hskip 0.1in
\includegraphics[width=0.22\textwidth]{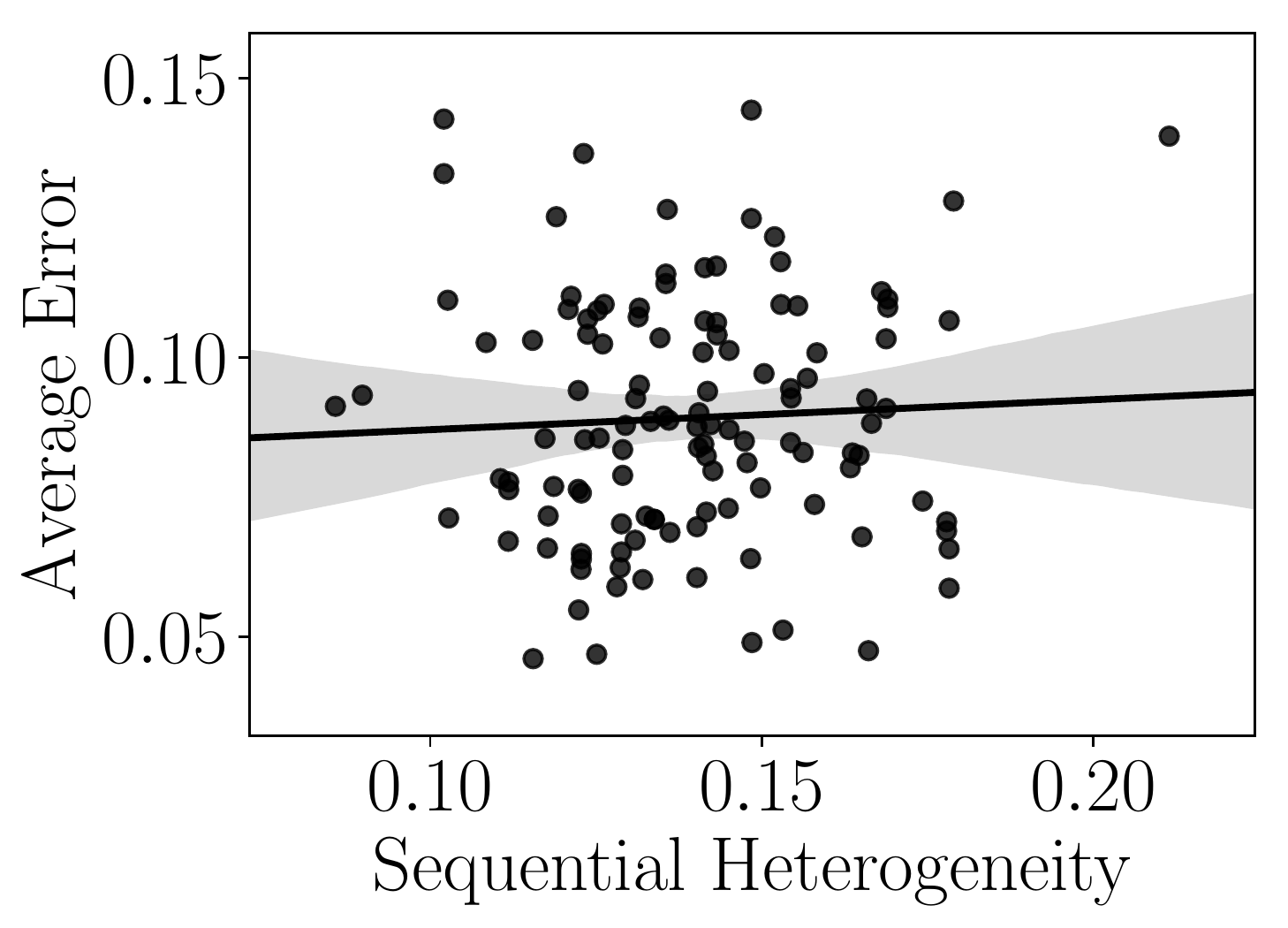}
\hskip 0.1in
\includegraphics[width=0.22\textwidth]{figs/hete_cifar_vcl.pdf}
\\
\hskip -0.24\textwidth
\raisebox{0.5in}{\rotatebox[origin=t]{90}{coreset VCL}}
\hskip 0.1in
\includegraphics[width=0.22\textwidth]{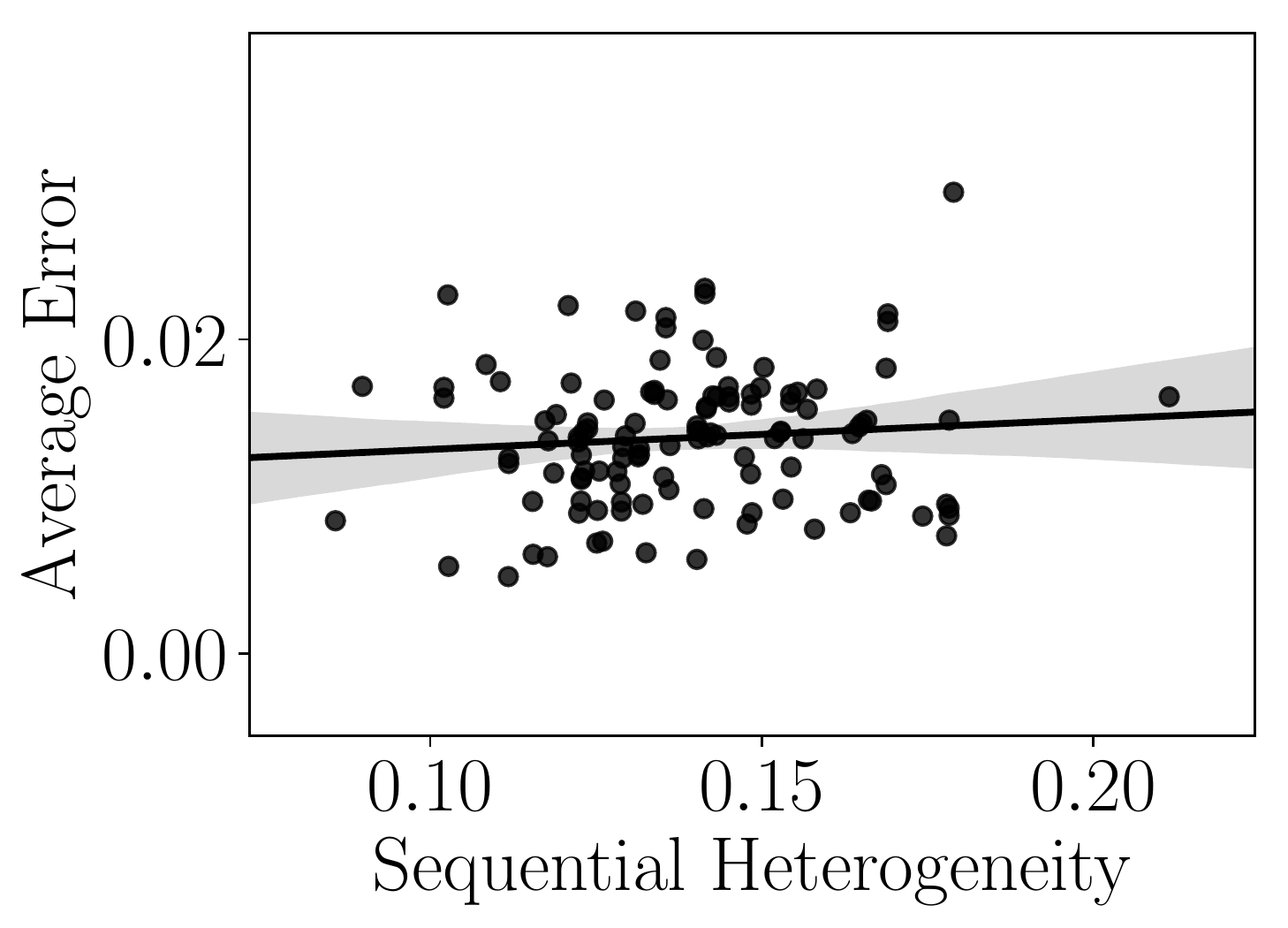}
\hskip 0.1in
\includegraphics[width=0.22\textwidth]{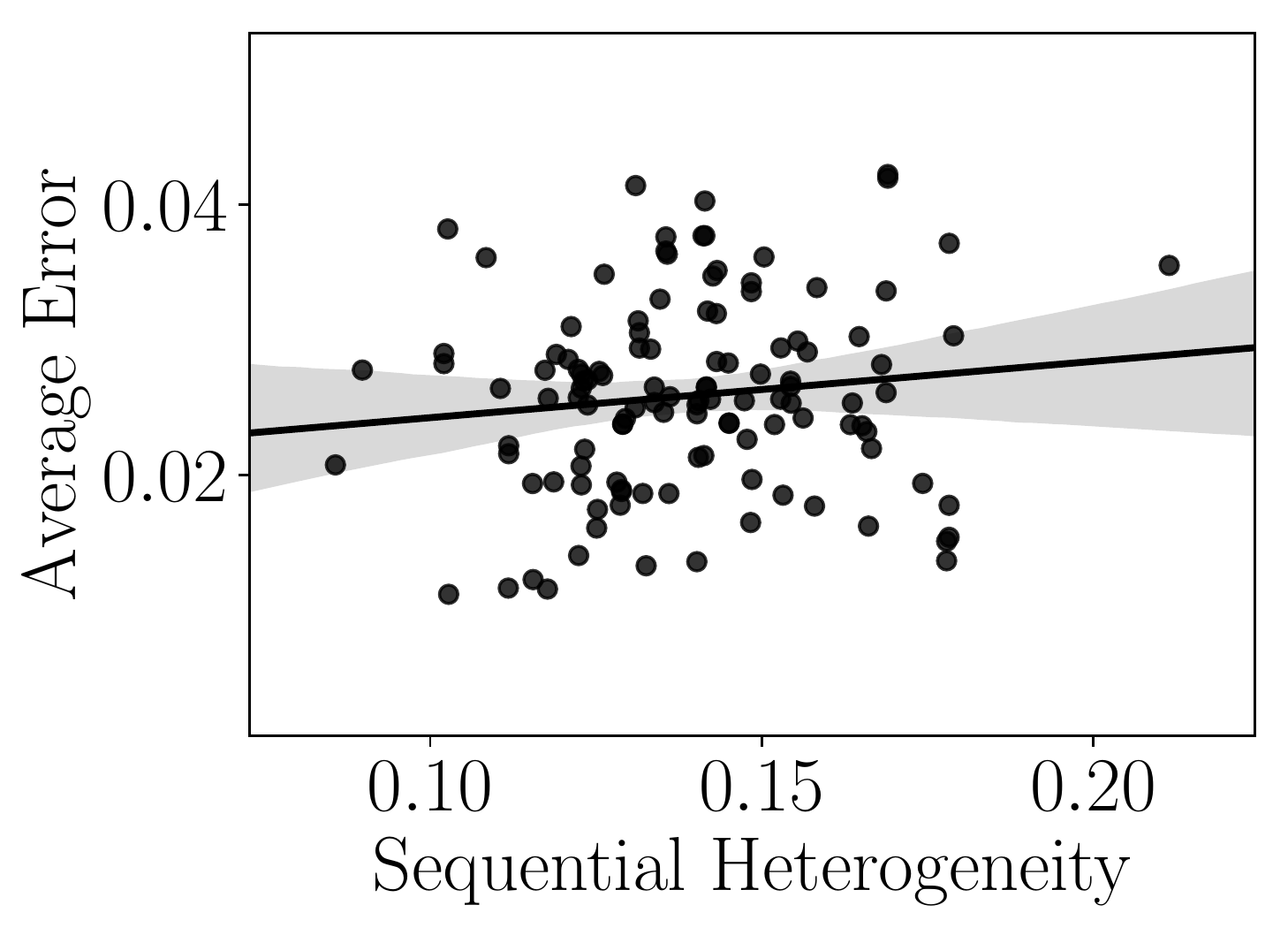}
\hskip 0.1in
\includegraphics[width=0.22\textwidth]{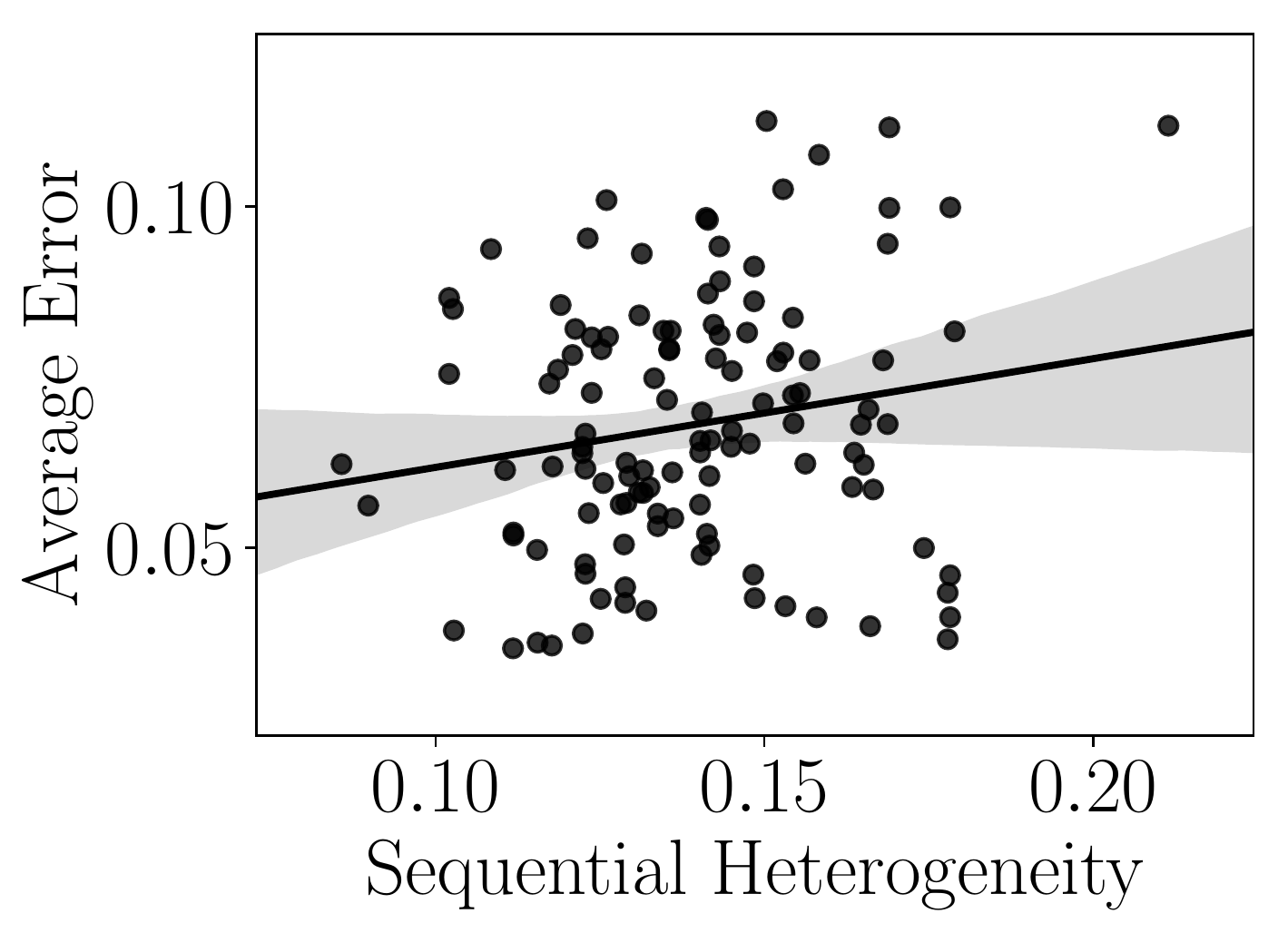}
\vskip 0.1in
\caption{{\bf Sequential heterogeneity vs. average error}, together with the linear regression fit and 95\% confidence interval, for each algorithm and test in Table 1(b). Green color indicates statistically significant positive correlations. Black color indicates negligible correlations.}
\label{fig:hete}
\end{center}
\end{figure*}

\begin{figure*}[!t]
\begin{center}
\raisebox{0.5in}{\rotatebox[origin=t]{90}{SI}}
\hskip 0.1in
\stackon[2pt]{\includegraphics[width=0.22\textwidth]{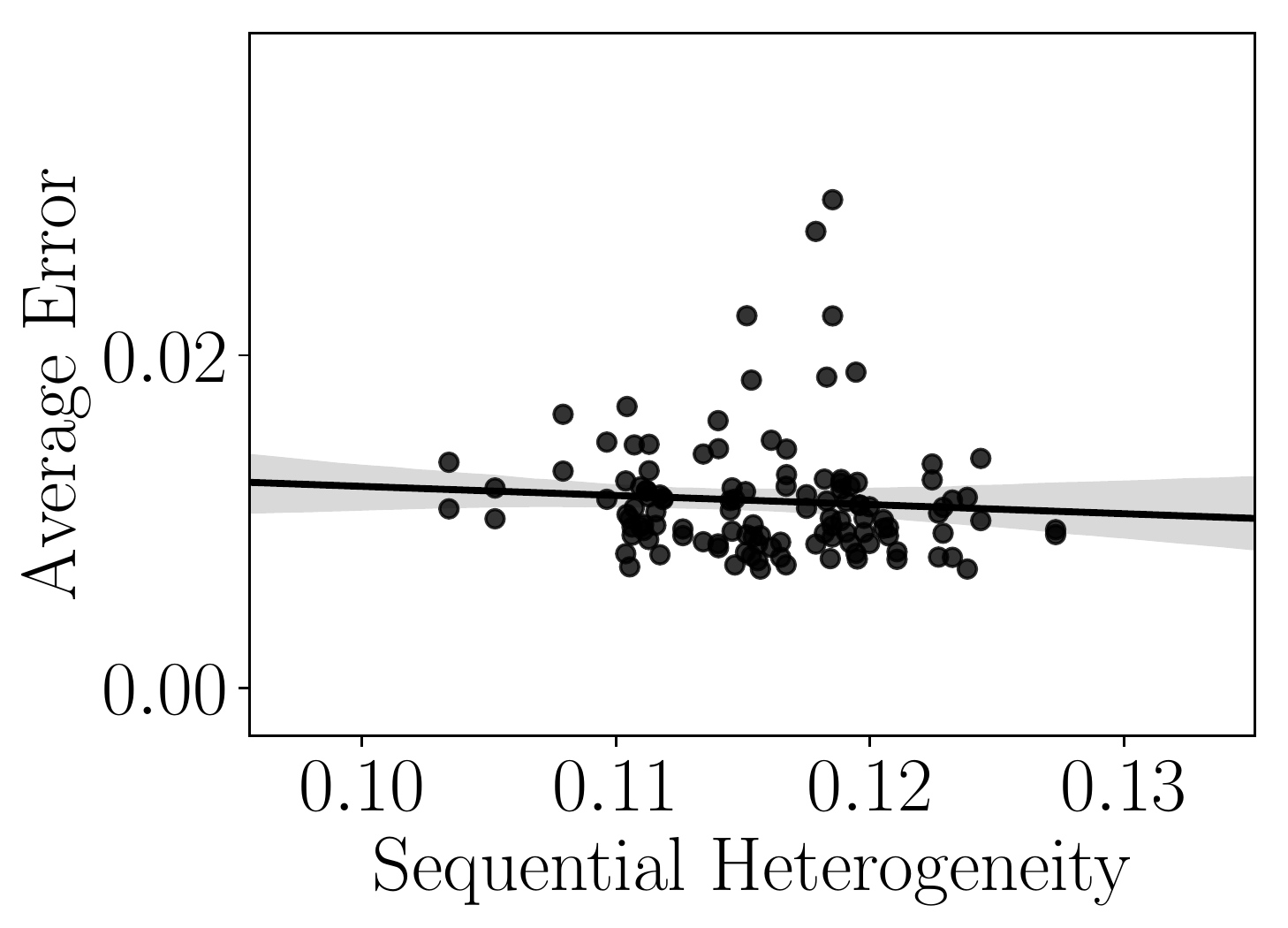}}{\quad MNIST-$256^2$}
\hskip 0.1in
\stackon[2pt]{\includegraphics[width=0.22\textwidth]{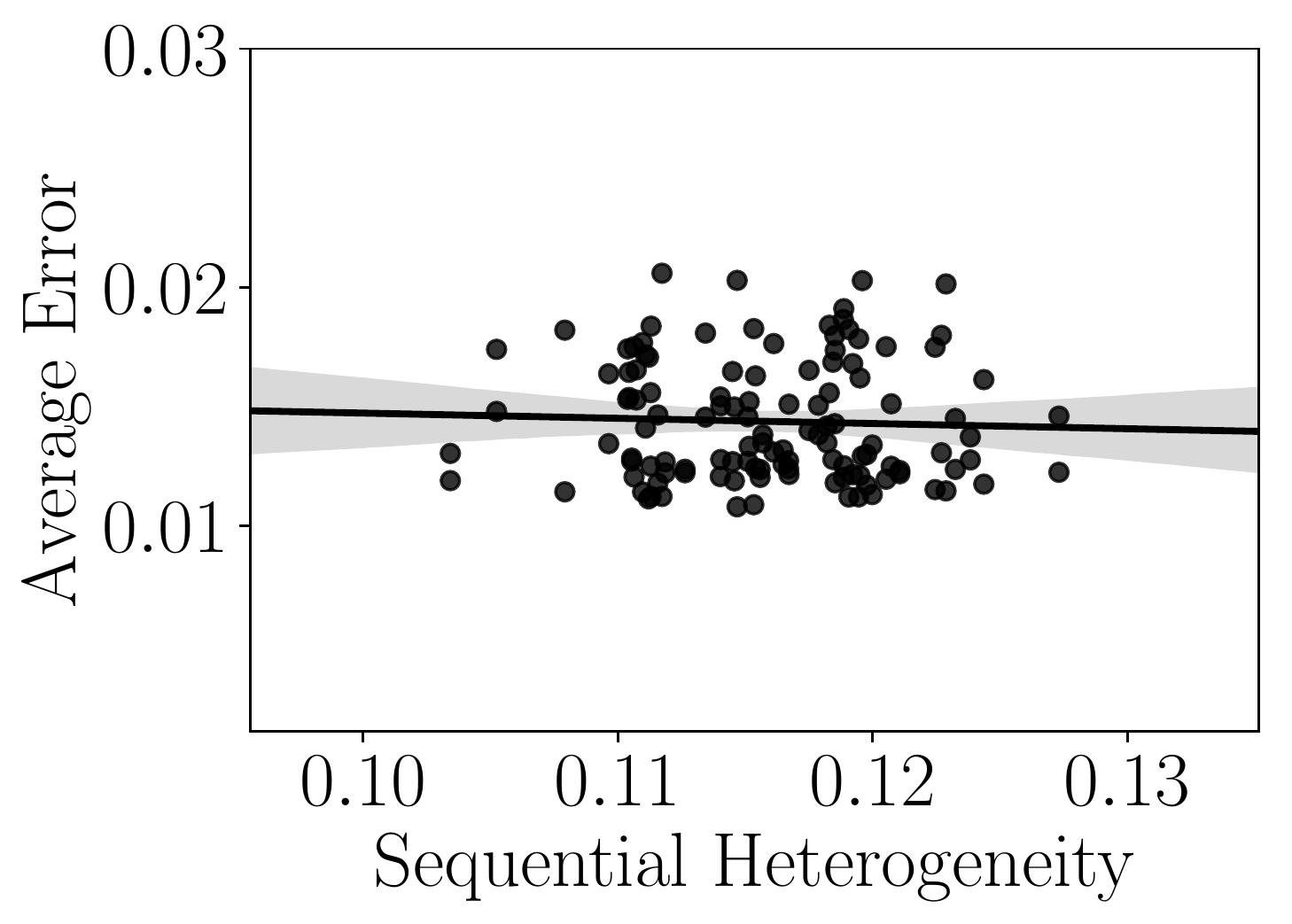}}{\quad MNIST-$50$}
\hskip 0.1in
\stackon[2pt]{\includegraphics[width=0.22\textwidth]{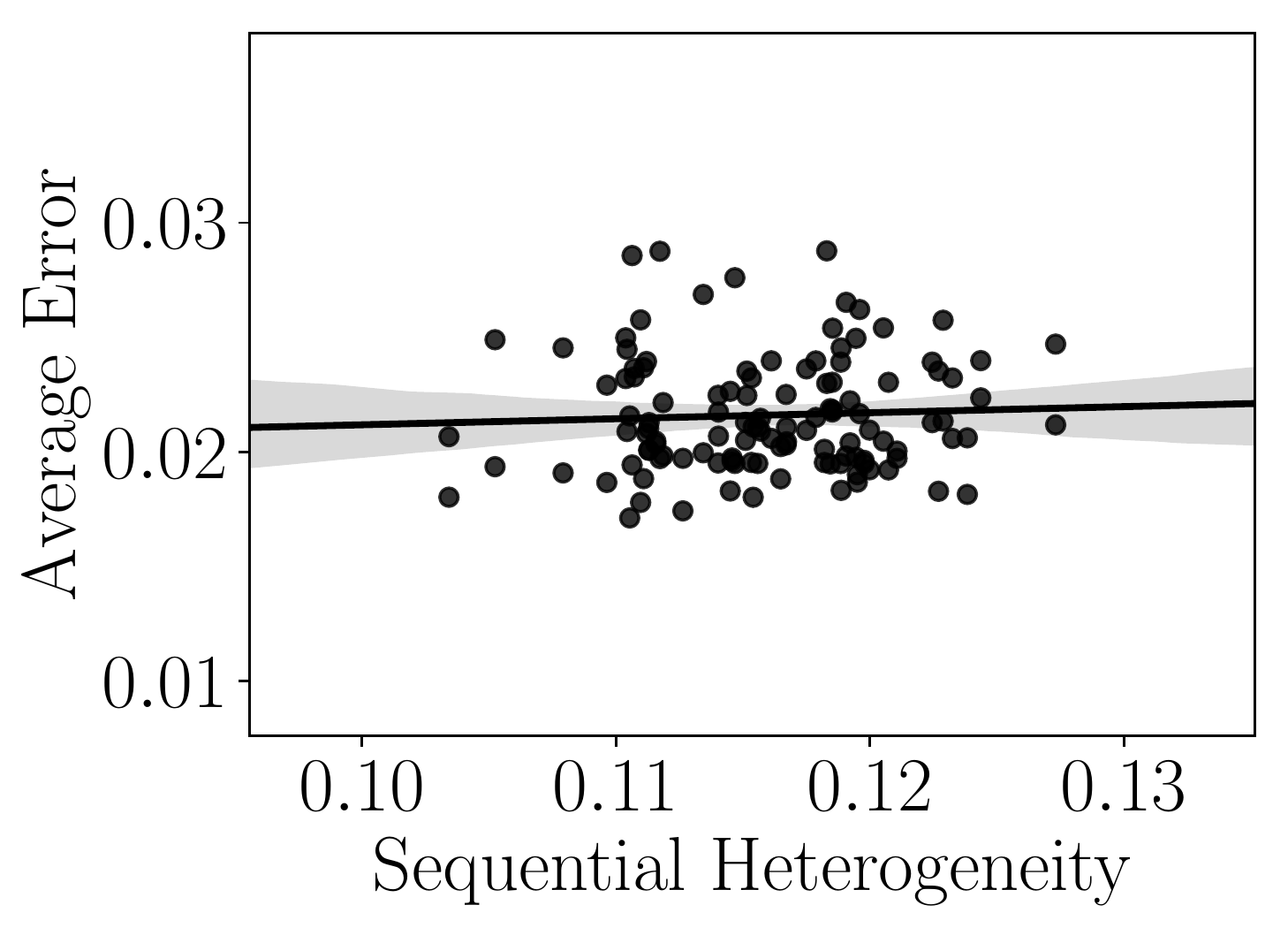}}{\quad MNIST-$20$}
\hskip 0.1in
\stackon[2pt]{\includegraphics[width=0.22\textwidth]{figs/perm_cifar_si.pdf}}{\quad CIFAR-10}
\\
\raisebox{0.5in}{\rotatebox[origin=t]{90}{VCL}}
\hskip 0.1in
\includegraphics[width=0.22\textwidth]{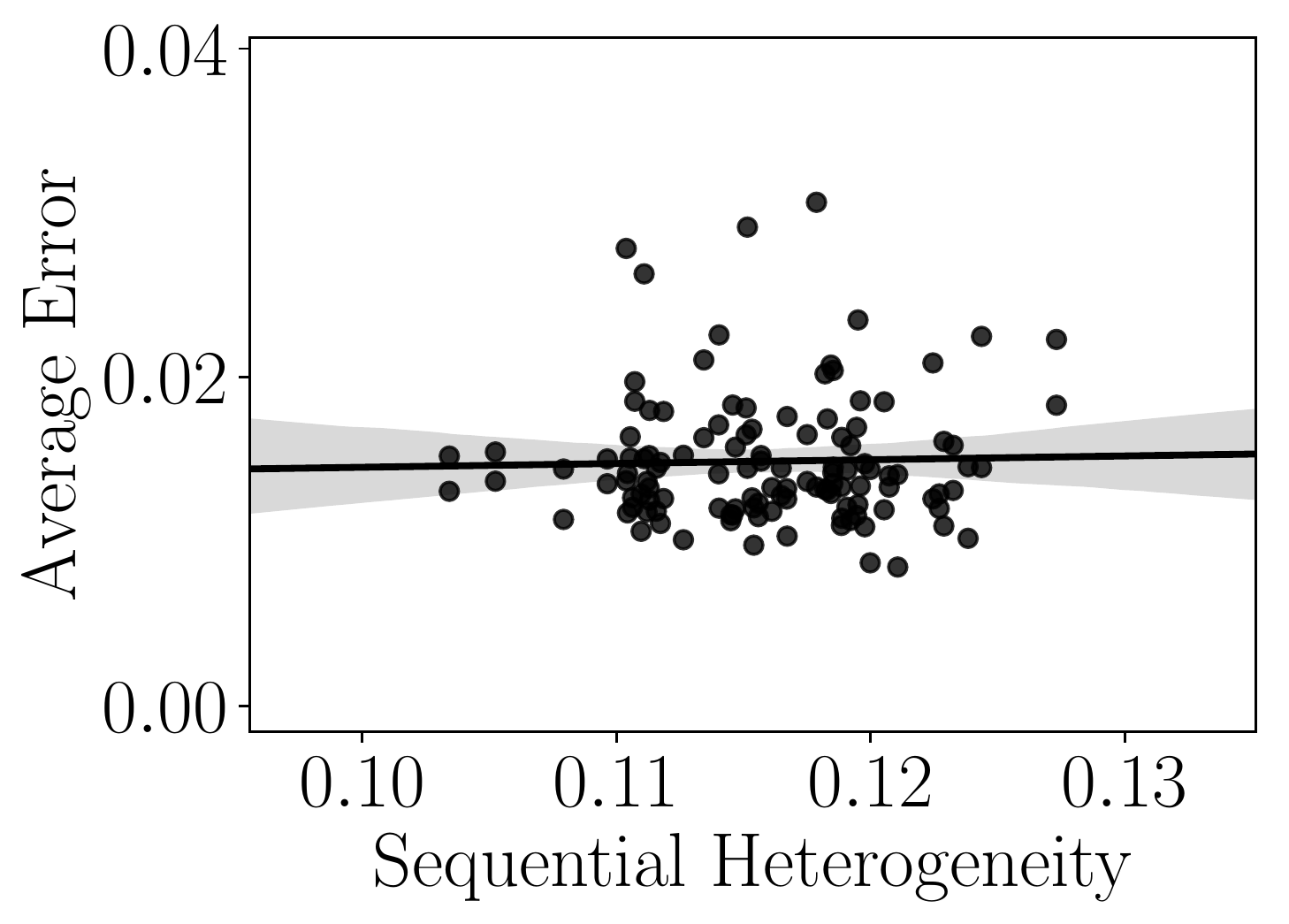}
\hskip 0.1in
\includegraphics[width=0.22\textwidth]{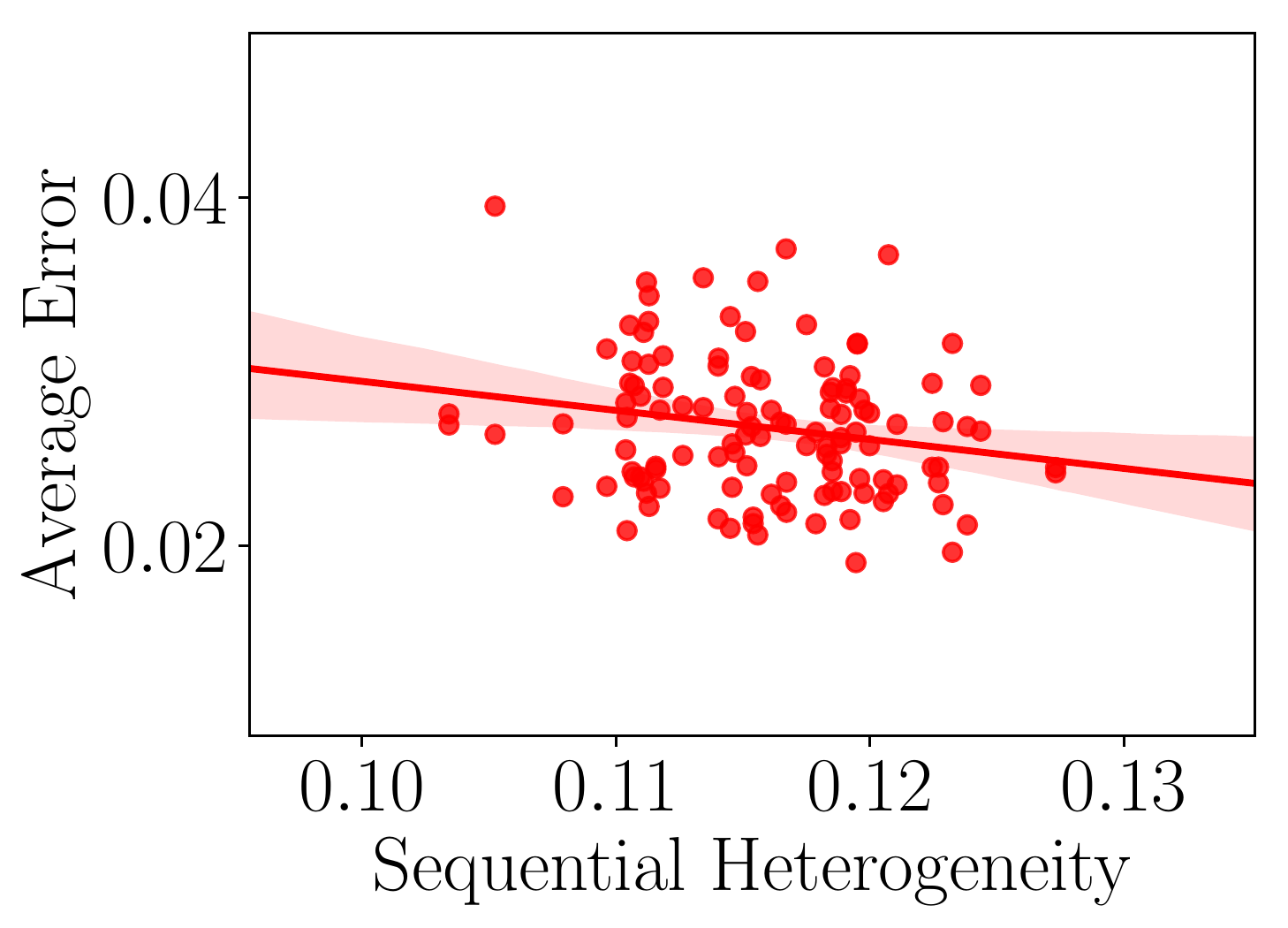}
\hskip 0.1in
\includegraphics[width=0.22\textwidth]{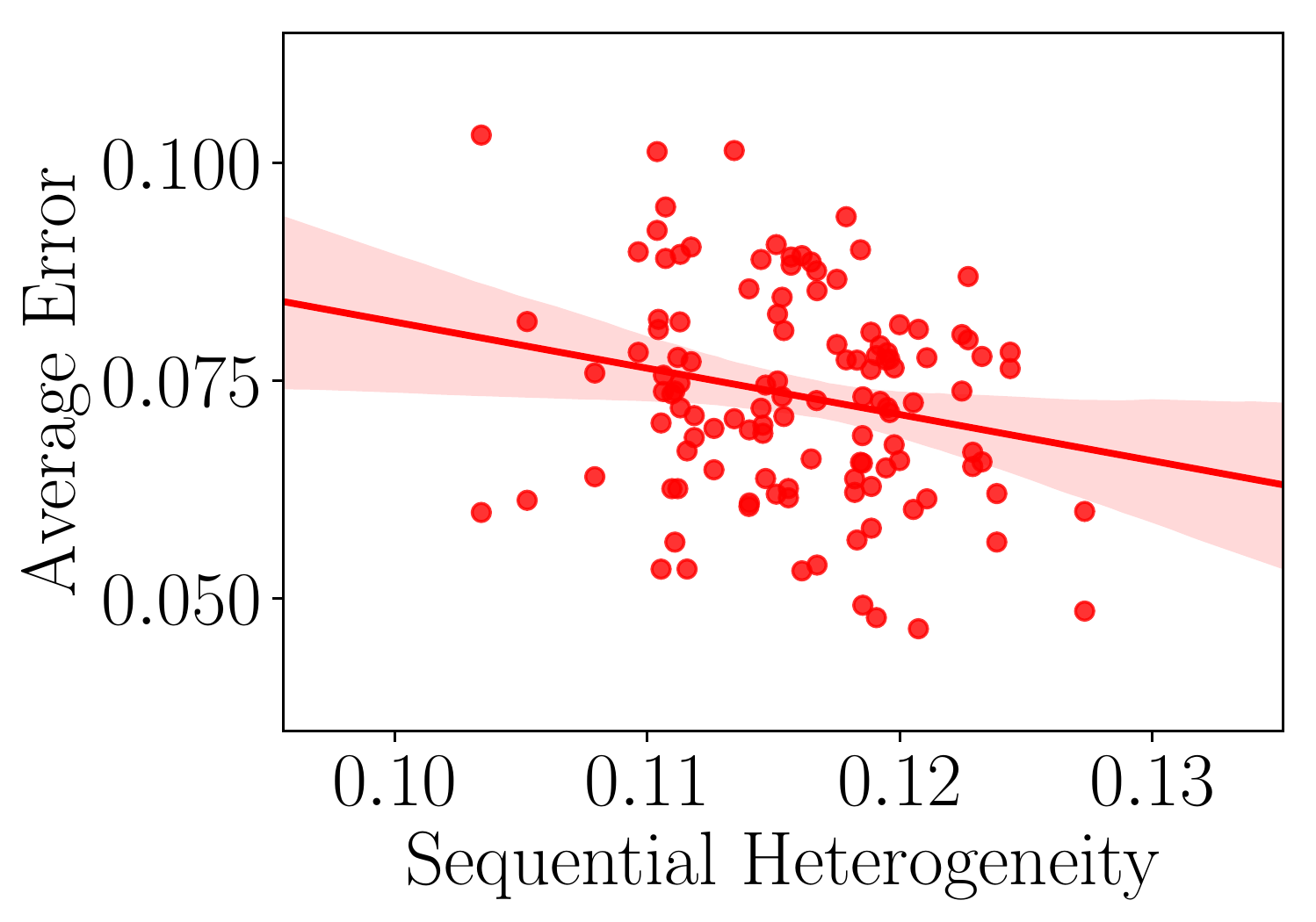}
\hskip 0.1in
\includegraphics[width=0.22\textwidth]{figs/perm_cifar_vcl.pdf}
\\
\hskip -0.24\textwidth
\raisebox{0.5in}{\rotatebox[origin=t]{90}{coreset VCL}}
\hskip 0.1in
\includegraphics[width=0.22\textwidth]{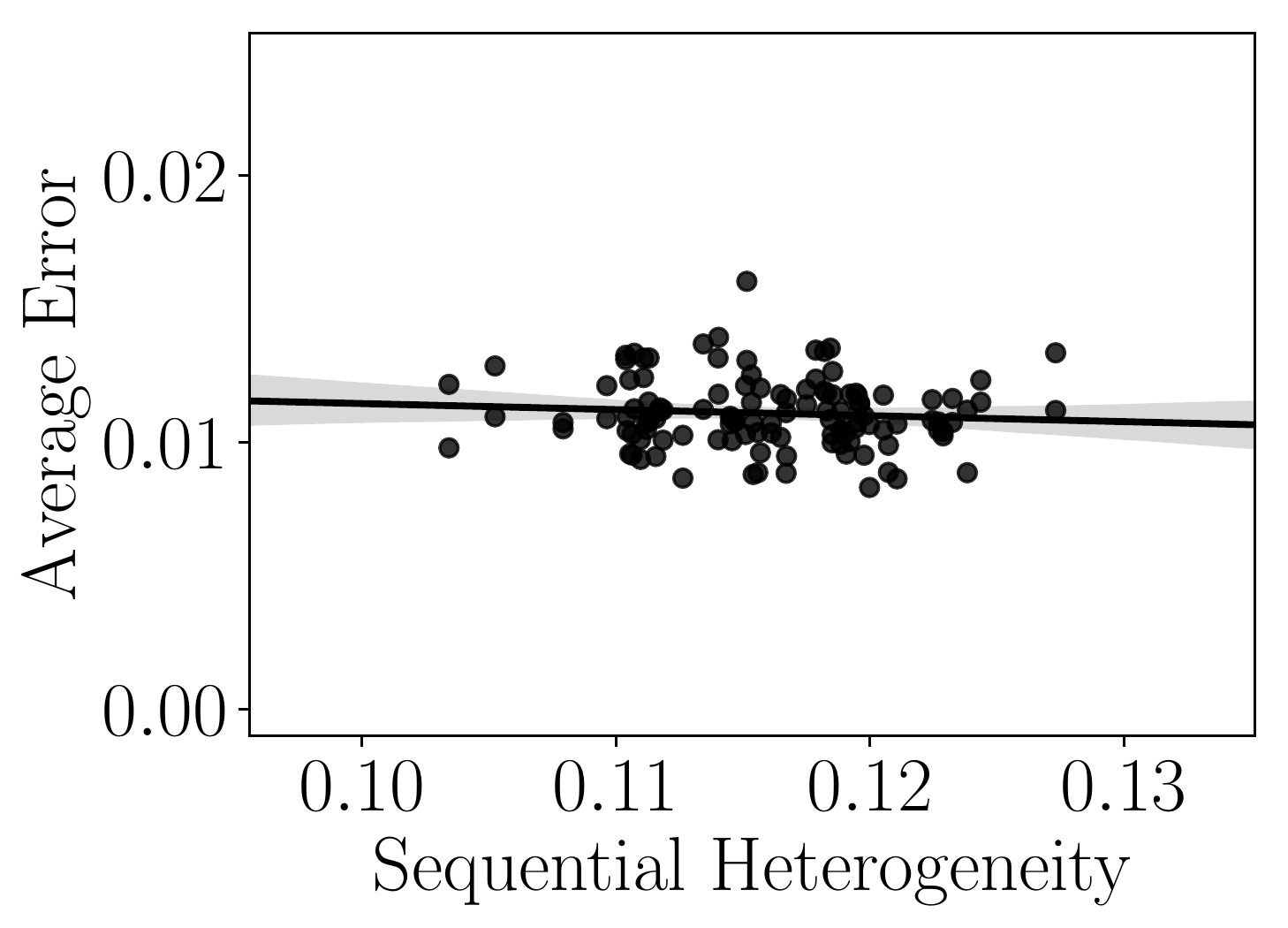}
\hskip 0.1in
\includegraphics[width=0.22\textwidth]{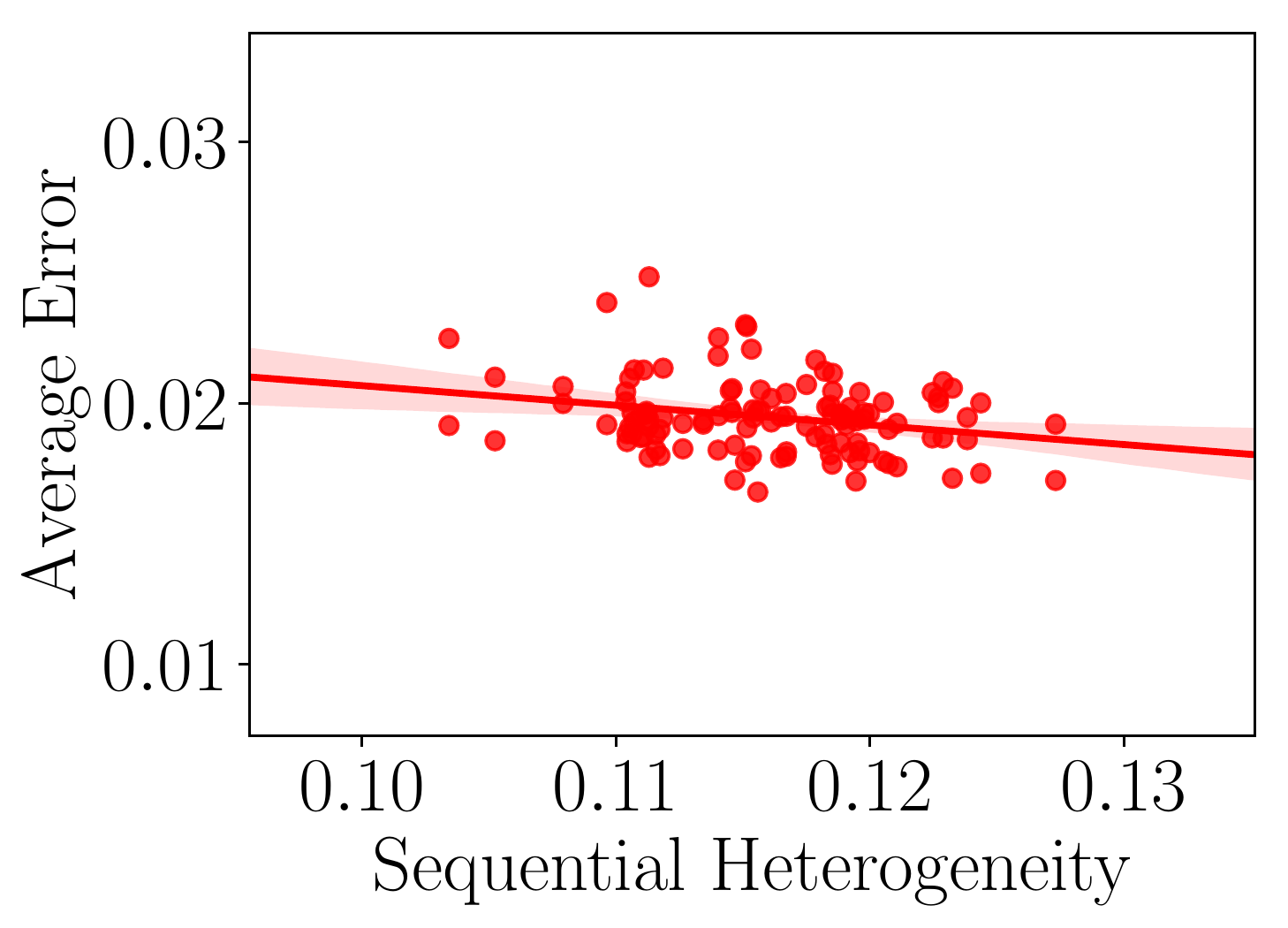}
\hskip 0.1in
\includegraphics[width=0.22\textwidth]{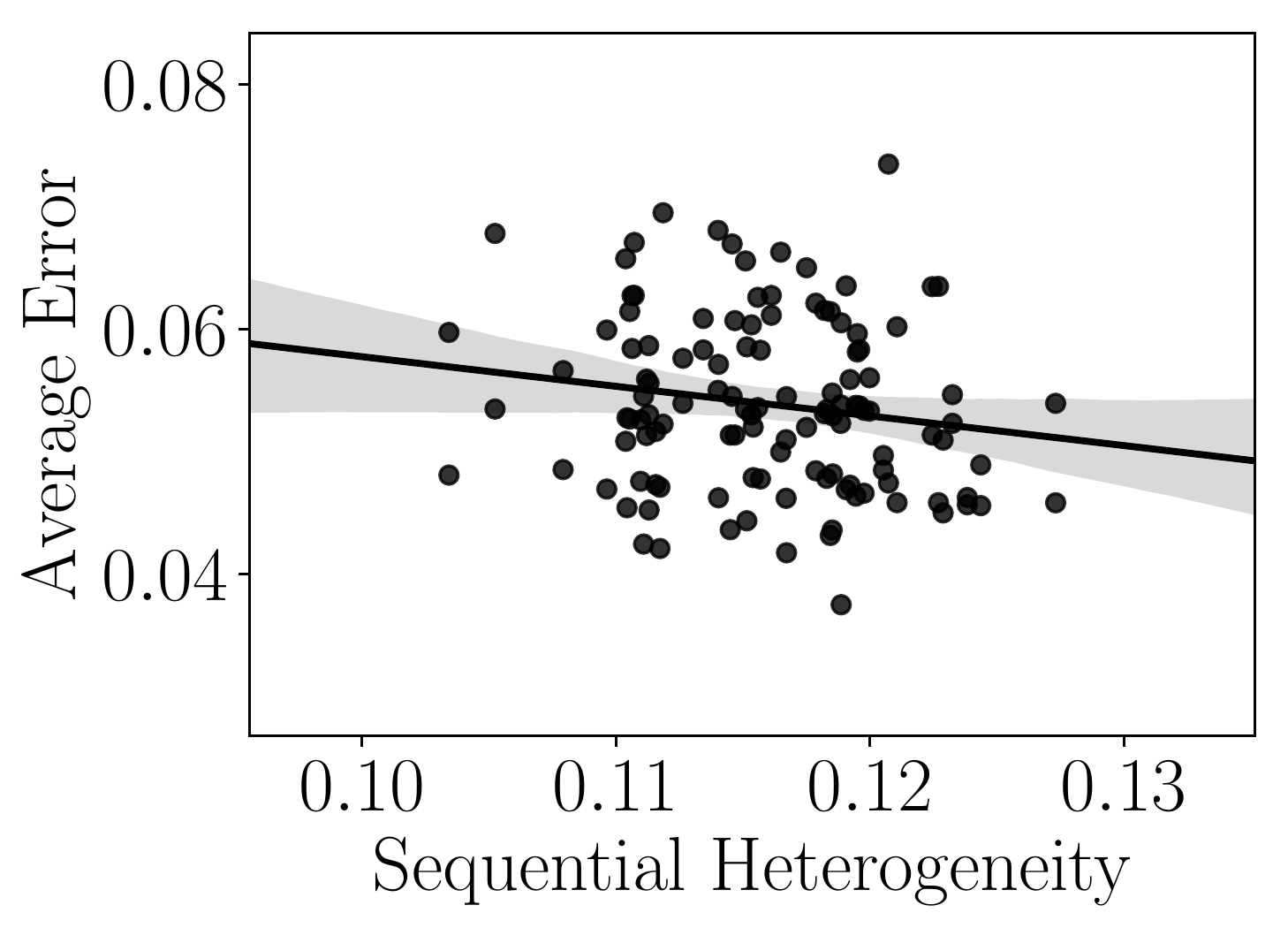}
\vskip 0.1in
\caption{{\bf Normalized sequential heterogeneity vs. average error}, together with the linear regression fit and 95\% confidence interval, for each algorithm and test in Table 1(c). Red color indicates statistically significant negative correlations. Black color indicates negligible correlations.}
\label{fig:norm-hete}
\end{center}
\end{figure*}

Tables~\ref{tab:result}(a--c) show the correlation coefficients and their p-values obtained from our experiments for the total complexity, sequential heterogeneity, and normalized sequential heterogeneity, respectively. We also show the scatter plots of the errors versus these quantities, together with the linear regression fits for the CIFAR-10 dataset in Fig.~\ref{fig:cifar}. All plots in the experiments, including those for the MNIST dataset, are provided in Fig.~\ref{fig:complexity},~\ref{fig:hete}, and~\ref{fig:norm-hete}.

Table~\ref{tab:result}(a) and Fig.~\ref{fig:cifar}(a) show strong positive correlations between error rate and total complexity for both SI and VCL in the CIFAR-10 setting, with a correlation coefficient of 0.86 for the former algorithm and 0.69 for the latter. These correlations are both statistically significant with p-values less than 0.01. On the MNIST-$256^2$ settings, SI and coreset VCL have weak positive correlations with total complexity, where the algorithms have correlation coefficients of 0.24 and 0.28, both with p-values less than 0.01, respectively.

When we reduce the capacity of the network and make the problem relatively harder (i.e., in the MNIST-$50$ and MNIST-$20$ settings), we observe stronger correlations for all three algorithms. With the smallest network (in MNIST-$20$), all the algorithms have statistically significant positive correlation with total complexity.

In terms of sequential heterogeneity, Table~\ref{tab:result}(b) and Fig.~\ref{fig:cifar}(b) show that it has a weak positive correlation with error rate in the CIFAR-10 setting. In particular, SI and VCL have correlation coefficients of 0.30 and 0.21 (both statistically significant), respectively. Interestingly, we find no significant correlation between error rate and sequential heterogeneity in all the MNIST settings, which suggests that heterogeneity may not be a significant factor determining the performance of continual learning algorithms on this dataset.

Since the complexity of each individual task in a sequence may influence the heterogeneity between the tasks (e.g., an easy task may be more similar to another easy task than to a hard task), the complexity may indirectly affect the results in Table~\ref{tab:result}(b). To avoid this problem, we also look at the normalized sequential heterogeneity in Table~\ref{tab:result}(c) and Fig~\ref{fig:cifar}(c), where the set of tasks is fixed and thus task complexity has been factored out. 

Surprisingly, Table~\ref{tab:result}(c) reports some negative correlations between error rate and sequential heterogeneity. For example, the correlation coefficient for SI on CIFAR-10 is -0.25 with a p-value less than 0.01, while there is no significant correlation for this algorithm on the MNIST dataset. VCL, on the other hand, has negative correlations with coefficients -0.20 and -0.21, respectively on MNIST-$50$ and MNIST-$20$, with p-values less than 0.05. Coreset VCL also has negative correlation between its error rate and sequential heterogeneity on MNIST-$50$, with coefficient -0.26 and p-value less than 0.01. These unexpected results suggest that in some cases, dissimilarity between tasks may even help continual learning algorithms, a fact contrary to the common assumption that the performance of continual learning algorithms would degrade if the tasks they need to solve are very different~\cite{ammar2014automated, ruder2017learning}.

\section{Discussions}
\label{sec:discuss}

\begin{figure*}[!t]
\begin{center}
\vskip -0.2in
\includegraphics[width=\textwidth]{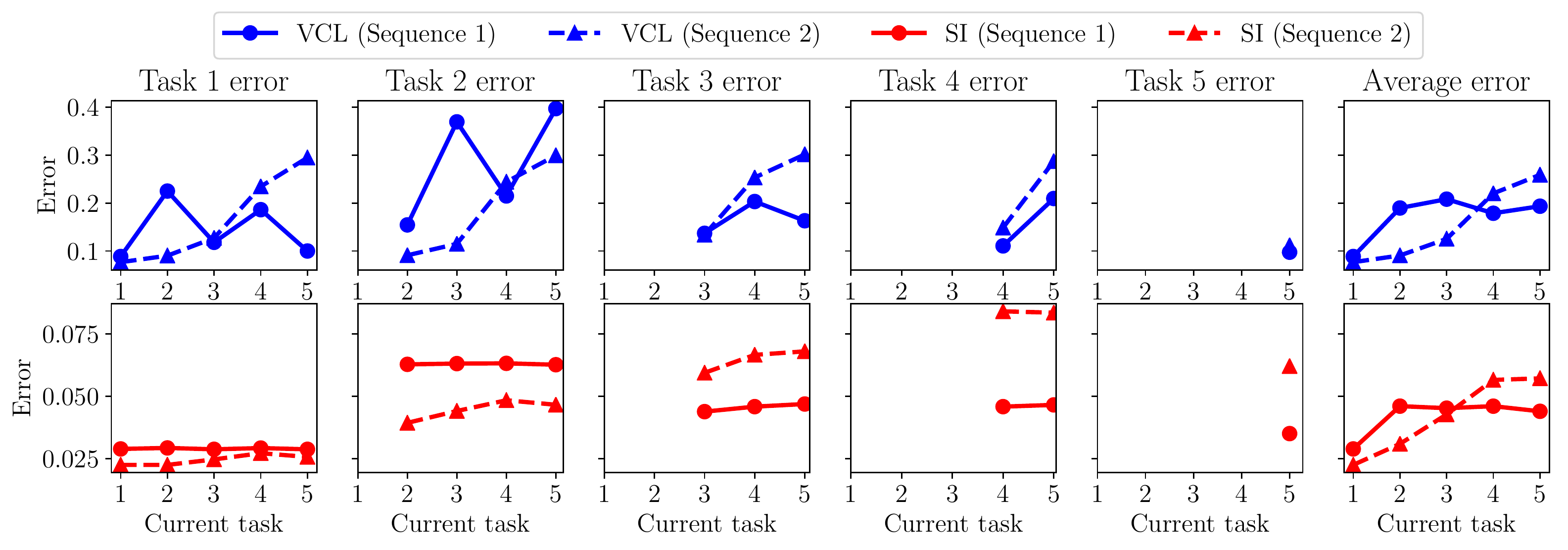}
\caption{{\bf Details of the error rates} of VCL and SI on two typical task sequences from CIFAR-10. Each column shows the errors on a particular task when subsequent tasks are continuously observed. Sequence 1 contains the binary tasks 2/9, 0/4, 3/9, 4/8, 1/2 with sequential heterogeneity 0.091, while sequence 2 contains the tasks 1/2, 2/9, 3/9, 0/4, 4/8 with sequential heterogeneity 0.068 (the labels are encoded to 0, 1, \ldots, 9 as usually done for this dataset). For both algorithms, the final average errors (the last points in the right-most plots) on sequence 2 are higher than those on sequence 1, despite sequence 1's higher sequential heterogeneity.}
\label{fig:norm-hete-details}
\vspace{-0.2in}
\end{center}
\end{figure*}

\begin{figure}[t]
\begin{center}
\includegraphics[width=\linewidth]{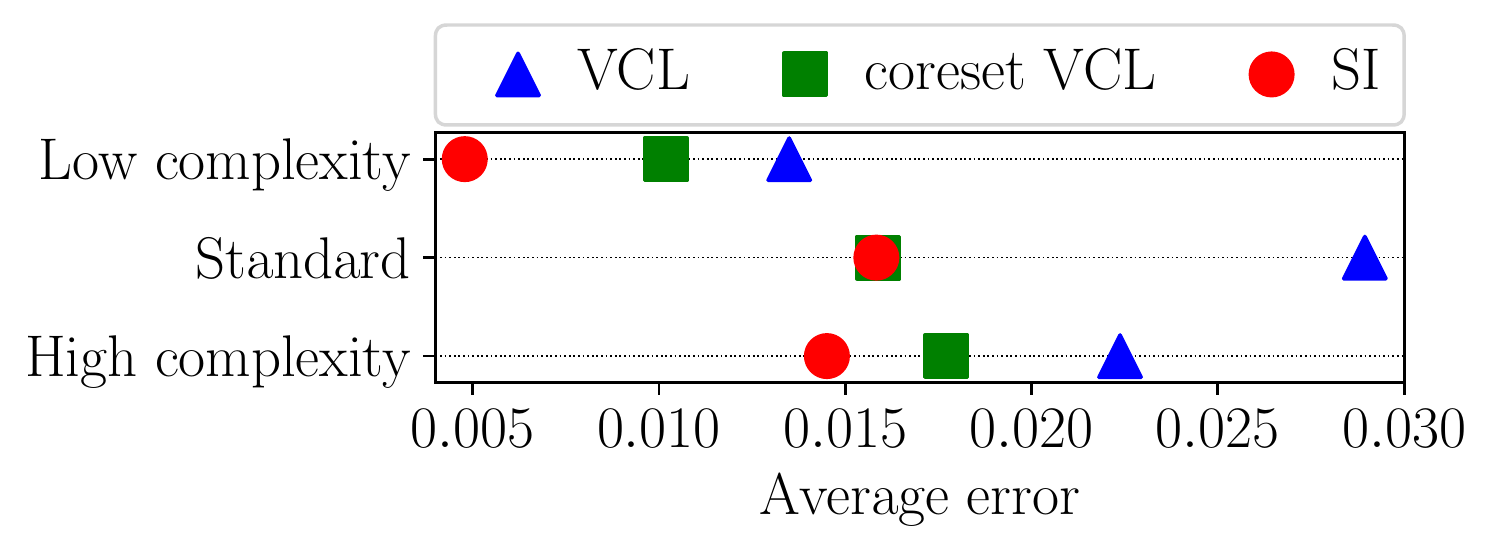}
\caption{{\bf Average error rates of VCL, coreset VCL and SI} on 3 task sequences from MNIST with different complexity levels. The high complexity sequence contains the binary tasks 0/1, 2/5, 3/5, 2/3, 2/6 with total complexity 0.48, while the low complexity sequence contains the tasks 0/1, 1/8, 1/3, 1/5, 7/8 with total complexity 0.35. The standard sequence contains the common split 0/1, 2/3, 4/5, 6/7, 8/9 with total complexity 0.41.}
\label{fig:mnist3}
\vspace{-0.2in}
\end{center}
\end{figure}

\minisection{On total complexity} 
The strong positive correlations between error rate and total complexity found in our analysis show that task complexity is an important factor in determining the effectiveness of continual learning algorithms. However, this factor is usually not taken into consideration when designing new algorithms or benchmarks. We suggest that task complexity is explicitly considered to improve algorithm and benchmark design. For example, different transfer methods can be used depending on whether one transfers from an easy task to a hard one or vice versa, rather than using a single transfer technique across all task complexities, as currently done in the literature. Similarly, when designing new benchmarks for continual learning, it is also useful to provide different complexity structures to test the effectiveness of continual learning algorithms on a broader range of scenarios and difficulty levels.

To illustrate the usefulness of comparing on various benchmarks, we construct two split MNIST sequences, one of which has high total complexity while the other has low total complexity. The sequences are constructed by starting with the binary classification task 0/1 and greedily adding tasks that have the highest (or lowest) complexity $C(t)$. Fig.~\ref{fig:mnist3} shows these sequences and the error rates of VCL, coreset VCL and SI when evaluated on them. We also show the error rates of the algorithms on the standard split MNIST sequence for comparison. From the figure, if we only compare on the standard sequence, we may conclude that coreset VCL and SI have the same performance. However, if we consider the other two sequences, we can see that SI is in fact slightly better than coreset VCL. This small experiment suggests that we should use various benchmarks, ideally with different levels of complexity, for better comparison of continual learning algorithms.

It is also worth noting that although the correlation between error rate and task complexity seems trivial, we are still not very clear which definition of task sequence complexity would be best to explain catastrophic forgetting (i.e., to give the best correlations). In this paper, we propose the first measure for this purpose, the total complexity.

\minisection{On sequential heterogeneity} 
The weak or negative correlations between error rate and sequential heterogeneity found in our analysis show an interesting contradiction to our intuition on the relationship between catastrophic forgetting and task dissimilarity. We emphasize that in our context, the weak and negative correlations are not a negative result, but actually a \emph{positive} result. In fact, some previous work showed that task similarity helps improve performance in the context of transfer learning~\cite{achille2019task2vec, ammar2014automated, ruder2017learning}, while some others claimed that task dissimilarity could help continual learning~\cite{farquhar2018towards} although their discussion was more related to the permuted MNIST setting. Our finding gives evidence that supports the latter view in the split MNIST and split CIFAR-10 settings.

To identify possible causes of this phenomenon, we carefully analyze the changes in error rates of VCL and SI on CIFAR-10 and observe some issues that may cause the negative correlations. For illustration, we show in Fig.~\ref{fig:norm-hete-details} the detailed error rates of these algorithms on two typical task sequences where the final average error rates do not conform with the sequential heterogeneity. Both of these sequences have the same total complexity, with the first sequence having higher sequential heterogeneity. 

From the changes in error rates of VCL in Fig.~\ref{fig:norm-hete-details}, we observe that for the first sequence, learning a new task would cause forgetting of its immediate predecessor task but could also help a task learned before that. For instance, learning task 3 and task 5 increases the errors on task 2 and task 4 respectively, but helps reduce errors on task 1 (i.e., backward transferring to task 1). This observation suggests that the dissimilarities between only consecutive tasks may not be enough to explain catastrophic forgetting, and thus we should take into account the dissimilarities between a task and all the previously learned tasks.

From the error rates of SI in Fig.~\ref{fig:norm-hete-details}, we observe a different situation. In this case, catastrophic forgetting is not severe, but the algorithm tends not to transfer very well on the second sequence. This inability to transfer leads to higher error rates on tasks 3, 4, and 5 even when the algorithm learns them for the first time. One possible cause of this problem could be that a fixed regularization strength $\lambda=1$ is used for all tasks, making the algorithm unable to adapt to new tasks well. This explanation suggests that we should customize the algorithm (e.g., by tuning the $\lambda$ values or the optimizer) for effectively transferring between different pairs of tasks in the sequence.

\minisection{Future directions}
The analysis offered by our paper provides a general and novel methodology to study the relationship between catastrophic forgetting and properties of task sequences. Although the two measures considered in our paper, total complexity and sequential heterogeneity, can explain some aspects of catastrophic forgetting, the correlations in Table~\ref{tab:result} are not very strong (i.e., their coefficients are not near 1 or -1). Thus, they can still be improved to provide better explanations for the phenomenon. Besides these two measures, we can also design other measures for properties such as intransigence~\cite{chaudhry2018riemannian}.

\section{Conclusion}
This paper developed a new analysis for studying relationships between catastrophic forgetting and properties of task sequences. An application of our analysis to two simple properties suggested that task complexity should be considered when designing new continual learning algorithms or benchmarks, and continual learning algorithms should be customized for specific transfers. Our analysis can be extended to study other relationships between algorithms and task structures such as the effectiveness of transfer or multi-task learning with respect to properties of tasks.

% \section*{Acknowledgements}

{\small
\bibliographystyle{ieee}
\bibliography{continual_task}
}

\end{document}